\pdfoutput=1

\documentclass[11pt]{article}

\usepackage[preprint]{acl}

\usepackage{times}
\usepackage{latexsym}

\usepackage[T1]{fontenc}

\usepackage[utf8]{inputenc}

\usepackage{microtype}

\usepackage{inconsolata}

\usepackage{mdframed}
\usepackage{graphicx}
\usepackage{float}
\usepackage{hyperref}
\usepackage{url}

\usepackage{amssymb}
\usepackage{amsmath}
\usepackage{epsfig,subfigure,caption}
\usepackage{algpseudocode}
\usepackage[normalem]{ulem}
\usepackage[linesnumbered,algoruled,boxed,noend]{algorithm2e}
\usepackage{xcolor}
\usepackage{color, colortbl}
\usepackage{multirow,booktabs, hhline}
\usepackage{bm}
\usepackage[linesnumbered,algoruled,boxed,noend]{algorithm2e}
\usepackage{wrapfig}
\usepackage{verbatimbox}
\usepackage{kantlipsum}
\usepackage{fancyvrb}
\usepackage[htt]{hyphenat}
\usepackage{pifont}
\newcommand{\cmark}{\ding{51}}%
\newcommand{\xmark}{\ding{55}}%

\usepackage[breakable]{tcolorbox}
\usepackage{enumitem}

\tcbset{
    myboxstyle/.style={
        colback=yellow!10!white, 
        colframe=yellow!50!black, 
        fonttitle=\bfseries,
        coltitle=black,
        boxrule=0.75mm,
        width=\textwidth,
        sharp corners,
        leftrule=1mm,
        left=5pt,
        right=5pt,
        top=5pt,
        bottom=5pt,
        fontupper=\small,  
        fontlower=\small   
    }
}
\newcommand{\dataset}[1]{{\textsc{RealTalk}}}

\newcommand{\secref}[1]{\S\ref{#1}}

\title{\dataset{}: A 21-Day Real-World Dataset for Long-Term Conversation}

\author{
    Dong-Ho Lee\textsuperscript{1}\thanks{~~Authors contributed equally. Work done during internship at Snap Inc.},~
    Adyasha Maharana\textsuperscript{2}$^*$,~
    \textbf{Jay Pujara\textsuperscript{1}},~
    \textbf{Xiang Ren\textsuperscript{1,3}},
    \textbf{Francesco Barbieri} \\
    \\
    \textsuperscript{1}University of Southern California~
    \textsuperscript{2}Databricks Mosaic Research~
    \textsuperscript{3}Sahara AI \\
    {\small 
        \texttt{\{dongho.lee, xiangren\}@usc.edu},
        \texttt{\{adyasha.maharana\}@databricks.com},
        \texttt{\{jpujara\}@isi.edu},~
        \texttt{\{fvancesco\}@gmail.com}
    }\\
}

\begin{document}
\maketitle
\begin{abstract}
Long-term, open-domain dialogue capabilities are essential for chatbots aiming to recall past interactions and demonstrate emotional intelligence (EI). 
Yet, most existing research relies on synthetic, LLM-generated data, leaving open questions about real-world conversational patterns.
To address this gap, we introduce \dataset{}, a 21-day corpus of authentic messaging app dialogues, providing a direct benchmark against genuine human interactions\footnote{\href{https://github.com/danny911kr/REALTALK}{https://github.com/danny911kr/REALTALK}}.

We first conduct a dataset analysis, focusing on EI attributes and persona consistency to understand the unique challenges posed by real-world dialogues.
By comparing with LLM-generated conversations, we highlight key differences, including diverse emotional expressions and variations in persona stability that synthetic dialogues often fail to capture.

Building on these insights, we introduce two benchmark tasks:
(1) \textbf{persona simulation} where a model continues a conversation on behalf of a specific user given prior dialogue context; and 
(2) \textbf{memory probing} where a model answers targeted questions requiring long-term memory of past interactions.

Our findings reveal that models struggle to simulate a user solely from dialogue history, while fine-tuning on specific user chats improves persona emulation.
Additionally, existing models face significant challenges in recalling and leveraging long-term context within real-world conversations.


\end{abstract}
\section{Introduction}


\begin{figure}[t!]
    \centering
    \begin{minipage}{\columnwidth}
    \centering
    \includegraphics[width=0.9\columnwidth]{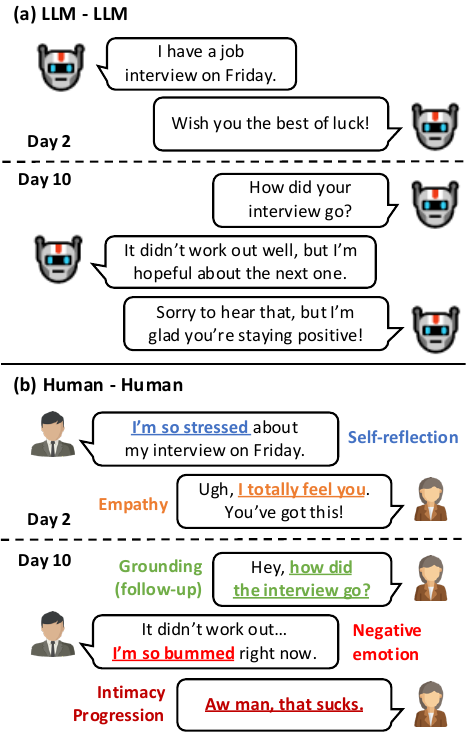}
\end{minipage}
    \vspace{-0.2cm}
    \caption{\textbf{A Motivation Example.} LLM-simulated dialogues often exhibit excessive empathy, even when discussing negative topics, whereas real-world human dialogues demonstrate a broader emotional spectrum, incorporate reflective and grounding language, and progressively develop intimacy over time.}
    \label{fig:motivation}
    \vspace{-0.5cm}
\end{figure}

Long-term open-domain dialogue capabilities are crucial for developing engaging chatbots that can remember previous interactions and respond with empathy, both of which are key aspects of emotional intelligence (EI)~\cite{goleman1998working, mayer2008emotional}.
However, evaluating large language models (LLMs) in this context is challenging.
A key obstacle lies in collecting \textbf{real-world, human-to-human} conversations featuring the \emph{same} individuals over extended periods, ensuring consistent personas—an important factor for passing the Turing test~\cite{turing2009computing, vinyals2015neural}.
To address this, recent efforts have turned to LLMs to simulate long-term dialogues between two human participants, resulting in synthetic conversation datasets~\cite{kim-etal-2023-soda, jang-etal-2023-conversation, zhong2024memorybank, du2024perltqa, maharana-etal-2024-evaluating}.
Yet, questions remain about whether these simulated dialogues genuinely capture the nuances of real-world human interaction.

In this study, we introduce \dataset{}, a real-world, long-term dialogue dataset featuring pairs of individuals who initially connected through messaging apps.
Over 21 days, 10 participants each engaged in two separate conversations with different partners, capturing evolving communication patterns. 
This resulted in a total of 10 unique conversations, each spanning approximately 21 sessions.
These daily or near-daily interactions amounted to over 16,000 words per conversation (\secref{ssec:data}).
The structure of these dialogues parallels the approach used in \textsc{LoCoMo}~\cite{maharana-etal-2024-evaluating}, where LLMs engage in extended conversations, exchanging small talk, and the occasional image.
By collecting human dialogues of comparable length, \dataset{} enables direct comparisons between real-world and LLM-simulated conversations.

Using our data, we analyze two key aspects of real-world long-term dialogues that distinguish them from LLM-simulated conversations.
First, we examine \textbf{emotional intelligence (EI)} by comparing authentic human conversations with LLM-generated dialogues using a dialogue-level EI assessment (\secref{sec:ei-evaluation} - \secref{ssec:analysis-speaker-level-ei}). 
Our findings show that real conversations gradually build intimacy over time~\cite{altman1973social, derks2008role}, featuring a wide emotional range, whereas LLM-simulated chats often begin at a high intimacy level, predominantly maintain a positive tone, and display excessive empathy—even amid negative topics (See Figure~\ref{fig:motivation}).
Next, we explore \textbf{persona consistency}, investigating whether individuals maintain a stable persona across multiple conversations and assessing how well models capture these shifts or continuities over time (\secref{ssec:analysis-persona}). 
Our analysis reveals that human conversational styles naturally fluctuate based on context, reflecting variations in emotional intelligence. 
In contrast, LLMs—despite being prompted with distinct personas—exhibit only minimal differences, struggling to adapt and replicate nuanced persona shifts over extended interactions.

Building on these insights, we introduce two benchmark tasks to evaluate model performance in long-term dialogues.  
(1) \textbf{Persona simulation}:
Maintaining a consistent persona is key to fostering trust, engagement, and personalization in AI interactions.  
This task assesses how well an LLM can simulate an individual’s unique persona by continuing a conversation a dialogue given prior context.
We evaluate the model’s ability to reproduce a user’s next message and match their EI attributes (\secref{ssec:persona-simulation});
(2) \textbf{Memory probing}:
Long-term memory is essential for AI to sustain coherent, context-aware conversations. 
This benchmark tests whether models can retain and apply long-term context by answering 728 human-annotated memory probing questions linked to the conversation (\secref{ssec:memory-probing}).  

Our experimental results show that models struggle to simulate a user solely from dialogue history. 
However, fine-tuning on a specific user’s chat history improves persona emulation. 
Additionally, existing models still face significant challenges in recalling and effectively leveraging long-term context in real-world conversations.

To summarize, we introduce \dataset{}, a large-scale, real-world dialogue dataset capturing 21 days of authentic, evolving conversations.  
Our analysis uncovers key differences in EI and persona consistency, showing that LLMs struggle to replicate human adaptability.
To bridge this gap, we introduce two benchmarks—persona simulation and memory probing—to drive the development of more human-like, memory-aware AI.

\begin{table*}[!th]
    \begin{minipage}{\textwidth}
    \centering
    \small
    \resizebox{\textwidth}{!}{
        \begin{tabular}{lcccccc}
            \toprule
            Dataset & Dialogue Participants & \# Turns / $\mathcal{C}$ & \# Session / $\mathcal{C}$ & \# Tokens / $\mathcal{C}$ & Multimodal & Collection \\
            \midrule
            MemoryBank~\cite{zhong2024memorybank} & Human-AI & 3.7 & 10 & 257.8 & \ding{55} & LLM-simulated \\
            LongMemEval~\cite{wu2024longmemeval} & Human-AI & 9.8 & 50.2 & 1,572.3 & \ding{55} & LLM-simulated \\

            \midrule
            SODA~\cite{kim-etal-2023-soda} & Human-Human & 7.6 & 1 & 122.4 & \ding{55} & LLM-simulated \\
            Conversation Chronicles~\cite{jang2023conversation} & Human-Human & 58.5 & 5 & 1,054.7 & \ding{55} & LLM-simulated \\
            LoCoMo~\cite{maharana-etal-2024-evaluating} & Human-Human & 588.2 & 27.2 & 13,377.2 & \ding{51} & LLM-simulated \\
            \midrule
            MPChat~\cite{ahn2023mpchat} & Human-Human & 2.8 & 1 & 53.3 & \ding{51} & Reddit \\
            MMDialog~\cite{feng2022mmdialog} & Human-Human & 4.6 & 1 & 72.5 & \ding{51} & Social media \\
            Daily Dialog~\cite{li2017dailydialog} & Human-Human & 7.9 & 1 & 114.7 & \ding{55} & Crowdsourcing \\
            MSC~\cite{xu2022beyond} & Human-Human & 53.3 & 4 & 1,225.9 & \ding{55} & Crowdsourcing \\
            \textbf{\dataset} & Human-Human & 894.4 & 21.9 & 17,109.8 & \ding{51} & Crowdsourcing \\
            \bottomrule
        \end{tabular}
    }
    \end{minipage}
    \vspace{-0.3cm}
    \caption{\textbf{Comparison of data statistics across various datasets.} \underline{Human-AI} dialogues primarily address task-oriented interactions where humans aim to achieve specific goals through the dialogue. 
    In contrast, \underline{Human-Human} dialogues involve conversational exchanges such as chit-chat or other forms of social interaction. 
    Unlike other works that simulate dialogues using models, \dataset{} is entirely derived from real-world human interactions.}
    \vspace{-0.5cm}
    \label{tab:data-statistics}
\end{table*}

\section{Related Work}
\paragraph{Long-term dialogues.}
Recent studies in long-term dialogue aim to improve model coherence and empathy through better memory recall of past interactions (See Table~\ref{tab:data-statistics} for more details).
Early work focused on collecting \textit{human-human} dialogues from online sources (\textit{e.g.,} Reddit) or via crowd-sourcing to evaluate model in multi-session, open-domain conversations~\cite{xu2022beyond, feng2022mmdialog, ahn2023mpchat, li2017dailydialog}, and collecting \textit{human-AI} dialogues to analyze user interactions and align models with user expectations~\cite{zheng2023lmsys, kopf2024openassistant, zhao2024wildchat}.
With LLMs now capable of processing longer contexts, recent studies simulate extensive \textit{human-human} dialogues~\cite{kim-etal-2023-soda, jang2023conversation, maharana-etal-2024-evaluating} and \textit{human-AI} dialogues~\cite{zhong2024memorybank, du2024perltqa, wu2024longmemeval} to improve model evaluation, addressing the difficulty of obtaining naturally long conversations.
While LLM simulations show some human-like traits, notable differences may remain compared to actual human interactions~\cite{hu-etal-2023-fine, kim-etal-2023-fantom, zhou2024real, mahowald2024dissociating, ivey2024real}.

In contrast to prior works, our work examines memory recalling capability of LLMs within real-world long-term \textit{human-human} dialogues, where informal cues may often challenge retention.

\paragraph{Emotional Intelligence.}
Prior research on emotional intelligence before the emergence of large language models (LLMs) primarily focused on affective understanding, where models were trained to recognize and interpret human emotions and sentiments based on text and its surrounding context~\cite{hu2004mining, pang-lee-2004-sentimental, hutto2014vader, rosenthal-etal-2017-semeval, mohammad-etal-2018-semeval, yin-etal-2020-sentibert, antypas2023supertweeteval}
With the advent of LLMs, researchers have adapted or fine-tuned these models for affective tasks, such as predicting emotion intensity and classifying emotions in human text~\cite{lei2023instructerc, liu2024emollms, zhang2023dialoguellm}.
Additionally, LLMs have been evaluated for their ability to infer how individuals might feel in specific scenarios~\cite{wang2023emotional, paech2023eq, huang2023emotionally, zhao2024both}.
However, existing studies remain limited to sentiment and emotion analysis at the message or context level, without assessing the overall emotional intelligence of a specific speaker across extended conversations.

In contrast, our work investigates long-term, real-world human dialogues, analyzing both genuine human interactions and LLM-simulated conversations through a comprehensive taxonomy of EI attributes. 
Beyond sentiment and emotion, we examine factors such as empathy, grounding acts, reflectiveness, and intimacy. Additionally, we use these EI attributes as metrics to evaluate how effectively a model simulates a specific persona over time.
\section{\dataset{}}
\label{ssec:data}

To study real-world human dialogues, we first collect \dataset{}, a dataset of 10 long-term conversations. 
We recruited 10 participants, pairing them to engage in conversations over 21 days.
Once the conversations were collected, a separate group of annotators, distinct from the participants, labeled memory probing question-answer pairs for each conversation $\mathcal{C}$ to evaluate LLM memory retention capabilities. 
Additionally, they annotated speaker events for each session $\mathcal{S}_i \in \mathcal{C}$ to track contextual information over time (\secref{ssec:subtask-annotation}).

\subsection{Data Collection}
\label{ssec:data-collection}
We provided specific guidelines to each participant, requiring them to send at least 50 messages to their conversation partner each day.
This section outlines the participant details, guidelines, and dataset statistics.

\paragraph{Participants.}
We recruited 10 participants who are native speakers from the US, aged between 18 and 25, with an approximately equal gender distribution. 
All participants provided signed release forms prior to the study.
For more details regarding the participants, see Appendix~\ref{appendix:chat-participant}

\paragraph{Guidelines.}
We asked participants to engage in friendly, natural conversations that include small talk, personal stories, and occasional image sharing to create a realistic chat experience. 
Conversations should feel casual and relatable, with references to time and place (e.g., ``last friday'' or ``when I was a kid'') and should cover topics such as daily events, personal interests, and pop culture.
Each session should feel like a conversation between friends, with participants sharing opinions and experiences.
Moreover, we encouraged participants to thoughtfully integrate images into the conversation, sourcing them from the internet under a Creative Commons license and avoiding explicit descriptions of the image content. 
We also advised against using images with identifiable faces representing the participant directly and requested consistency if such images are reused across sessions. 
Detailed guidelines are provided in Appendix~\ref{appendix:chat-guidelines}, and dataset statistics are discussed in \secref{ssec:analysis-data}.

\subsection{Annotation for Memory Probing}
\label{ssec:subtask-annotation}
Memory probing is essential for evaluating an LLM’s ability to retain and retrieve information from long-term conversations, a key challenge in developing coherent, contextually aware AI systems.
we recruited a separate group of annotators to create memory-probing QA pairs and identify key events for each session $S_i$.
The detailed annotation guidelines provided to annotators are available in Appendix~\ref{appendix:qa-guidelines}.

\paragraph{QA Annotations}
annotators generated questions that require referencing prior interactions to be correctly answered, following the procedure from \citet{maharana-etal-2024-evaluating}. 
We categorize these questions into three types:
(1) \textsc{Multi-hop} requires synthesizing information across multiple sessions, represented as \( \sum_{i \in \mathcal{S}_{\text{selected}}} \mathcal{S}_i \), where \( \mathcal{S}_{\text{selected}} \) denotes the subset of sessions chosen by the annotator;
This evaluates a model’s ability to connect dispersed details;
(2) \textsc{Temporal reasoning} involves reasoning over the sequence of events and time-based dependencies, assessing whether a model can track evolving narratives and maintain temporal coherence; and
(3) \textsc{Commonsense} cannot be answered solely from the conversation and requires commonsense reasoning, testing a model’s ability to integrate contextual dialogue with external knowledge.
In total, 728 questions were annotated with answers, comprising 302 multi-hop, 321 temporal reasoning, and 111 commonsense questions (see details in Appendix~\ref{appendix:qa-statistics}).

\paragraph{Event Annotations}
For each session \( \mathcal{S}_i \in \mathcal{C} \), annotators are asked to document all events \( \mathcal{E}_j \in \mathcal{S}_i \) occurring in each speaker’s life in a free-text format. 
These events include those that have already happened, are planned for the future, or are ongoing during the conversation. 
Events can range from smaller occurrences, like taking a cooking class, to major life events, such as enrolling in college or traveling to another country.

\section{Emotional Intelligence (EI) Evaluation}
\label{sec:ei-evaluation}
To systematically compare real-world and LLM-generated dialogues, and to evaluate persona consistency and simulation quality, we develop a comprehensive EI evaluation framework.

\subsection{Problem formulation.}
Each conversation $\mathcal{C}$ consists of multiple sessions $\mathcal{S}_i$, such that $\mathcal{C} = \{\mathcal{S}_1, \mathcal{S}_2, \dots, \mathcal{S}_m\}$, where $m$ is the total number of sessions in the conversation.
Each session $\mathcal{S}_i$ consists of a sequence of messages exchanged between two speakers such that $\mathcal{S}_i = \{\mathcal{M}_{s_1,1}, \mathcal{M}_{s_2,2}, \dots, \mathcal{M}_{s_n,n}\}$ where \( \mathcal{M}_{s_j,j} \) represents the \( j \)-th message in the session, sent by speaker \( s_j \in \{1,2\} \).
Unlike structured turn-taking, the order of messages can be variable, and a speaker may send consecutive messages (\textit{e.g.,} multiple chat bubbles in a row), reflecting the natural flow of real-world conversations.
To maintain coherence, we concatenate consecutive messages from the same speaker into a single message before analysis.

\subsection{Message-level EI Attributes.}
\label{ssec:message-level-ei}
We measure following EI attributes for each individual message $\mathcal{M}$:

\paragraph{Reflectiveness.} $R(\mathcal{M})$ is a boolean indicator that captures whether a speaker explicitly recognizes and describes their own emotions, thoughts, or reactions. 
For example, statements like ``I think I’m feeling this way because...'' signal self-reflection. 
We use an LLM-based classification to label each turn as reflective or not (see Appendix~\ref{appendix:evaluation-reflective}).

\paragraph{Grounding Act.}
$G(\mathcal{M})$ is a boolean indicator that captures whether a speaker use acts—such as clarifying questions or follow-ups—which are essential for building common ground and preventing misunderstandings~\cite{clark1996using, clark1989contributing, shaikh-etal-2024-grounding}.
We use an LLM-based classification to label each turn as grounding or not (see Appendix~\ref{appendix:evaluation-grounding}).

\paragraph{Emotion \& Sentiment \& Intimacy.}
$E(\mathcal{M})$ and $S(\mathcal{M})$ represent the speaker's emotion and sentiment labels in a message $\mathcal{M}$, respectively.
Emotion is classified into 11 predefined categories (e.g., anger, fear, joy), while sentiment is classified into 3 categories (positive, negative, neutral). 
$I(\mathcal{M})$ is a floating-point value that indicates the speaker's level of intimacy in the message $\mathcal{M}$.
Each of these attributes is measured using models trained specifically for emotion, sentiment, and intimacy classification, leveraging an annotated dataset from Twitter~\cite{antypas2023supertweeteval}.
While LLMs could be used for these tasks, fine-tuned models perform comparably or slightly better while being significantly more cost-efficient, making them a more practical choice for large-scale evaluation~\cite{rathje2024gpt}.

\paragraph{Empathy.}
$EP(\mathcal{M})$ is a floating-point score capturing the speaker's empathy in a message $\mathcal{M}$.
It consists of three components: emotional reaction, interpretation, and exploration. 
Each component is scored by the LLM on a 0–2 Likert scale based on the \textsc{EPITOME} framework~\cite{sharma-etal-2020-computational} (see Appendix~\ref{appendix:evaluation-empathy}). 
The empathy score for a message is the sum of the scores for these components.

\subsection{Speaker-level EI Evaluation.}
\label{ssec:speaker-level-ei}
To assess a speaker’s overall emotional intelligence (EI), we aggregate message-level EI attributes into five categories based on Goleman’s EI framework~\cite{goleman1998working}.

\paragraph{Self-awareness.}
It reflects a speaker’s ability to recognize and articulate emotions and perspectives. 
It is measured using two metrics: 

\vspace{0.2cm}\noindent
(1) \textit{Reflective frequency} measures how often a speaker engages in self-reflection:
\[
\text{reflective\_frequency}(s) = \frac{\sum_{\mathcal{M} \in \mathcal{M}_s} R(\mathcal{M})}{|\mathcal{M}_s|}
\]
where \(\mathcal{M}_s\) is the set of messages by speaker \(s\).

\vspace{0.2cm}\noindent
(2) \textit{Emotion \& Sentiment Diversity} captures the range of emotions or sentiments expressed by a speaker \(s\), calculated using entropy.
Let \(L\) represent the predefined sentiment labels (\textit{i.e.,} positive, negative, neutral), and \(p_s(x)\) denote the proportion of label \(x \in L\) in the speaker's messages \(\mathcal{M}_s\):
\[
p_s(x) = \frac{\sum_{\mathcal{M} \in \mathcal{M}_s} 1 (S(\mathcal{M}) = x)}{|\mathcal{M}_s|},
\]
where \(1 (S(\mathcal{M}) = x)\) equals 1 if the sentiment label of message \(\mathcal{M}\) is \(x\), and 0 otherwise.  
Using this, sentiment diversity is computed as:
\[
\text{sentiment\_diversity}(s) = - \sum_{x \in L} p_s(x) \cdot \log_2 p_s(x).
\]
The same applies to \textit{emotion\_diversity} by substituting \(S(\mathcal{M})\) with \(E(\mathcal{M})\) and using emotion labels.


\paragraph{Motivation}
A speaker’s engagement in maintaining the conversation, measured by \textit{grounding frequency}, the proportion of messages containing grounding acts:  
\[
\text{grounding\_frequency}(s) = \frac{\sum_{\mathcal{M} \in \mathcal{M}_s} G(\mathcal{M})}{|\mathcal{M}_s|}
\]

\paragraph{Social Skills}
A speaker’s ability to foster trust and engagement, measured via \textit{intimacy progression}, assuming intimacy naturally evolves over time.
The average intimacy for speaker $s$ per session $\mathcal{S}_i$ is:
\[
\text{intimacy\_average}(s, \mathcal{S}_i) = \frac{\sum_{\mathcal{M} \in \mathcal{M}_s^{\mathcal{S}_i}} I(\mathcal{M})}{|\mathcal{M}_s^{\mathcal{S}_i}|}
\] 
where \( \mathcal{M}_s^{\mathcal{S}_i} \) is the set of messages from speaker \( s \) in session \( \mathcal{S}_i \), and \( I(\mathcal{M}) \) denotes the intimacy score of message \( \mathcal{M} \).
The progression of average intimacy across sessions in \(\mathcal{C}\) is modeled as:  
\[
\text{linear\_progression}(s) = a, \text{ where } y = a \cdot x + b
\]  
\[
\text{exp\_progression}(s) = b, \text{ where } y = a \cdot e^{b \cdot x}
\]

\paragraph{Self-regulation}
It reflects a speaker’s ability to maintain emotional and sentiment stability while interacting with their conversational partner.
We evaluate this using two metrics:

\vspace{0.2cm}\noindent
(1) \textit{Emotion \& Sentiment Stability} measures the consistency of a speaker's emotions or sentiments across their messages:
\[
\text{stability}(s) = \frac{\sum_{k=2}^{|\mathcal{M}_s|} 1(E(\mathcal{M}_s^{(k)}) = E(\mathcal{M}_s^{(k-1)}))}{|\mathcal{M}_s| - 1}
\]  
where \( \mathcal{M}_s^{(k)} \) is the \( k \)-th message in speaker \( s \)'s ordered sequence of messages, \( 1(\cdot) \) is an indicator function that returns 1 if the emotion of \( \mathcal{M}_s^{(k)} \) matches the previous message \( \mathcal{M}_s^{(k-1)} \), and 0 otherwise.
The same formula applies for sentiment stability, replacing \( E(\mathcal{M}) \) with \( S(\mathcal{M}) \).  

\vspace{0.2cm}\noindent
(2) \textit{Emotion \& Sentiment Alignment} measures a speaker's synchronization with a partner:  
\[
\text{alignment}(s) = \frac{\sum_{k} 1(E(\mathcal{M}_s^{(k)}) = E(\mathcal{M}_p^{(k)}))}{|\mathcal{M}_s|}
\]  
where \( \mathcal{M}_s^{(k)} \) and \( \mathcal{M}_p^{(k)} \) are corresponding messages from speaker \( s \) and their partner \( p \), and \( 1(\cdot) \) returns 1 if their emotions match and 0 otherwise.  
The same formula applies for sentiment alignment, substituting \( E(\mathcal{M}) \) with \( S(\mathcal{M}) \).

\paragraph{Empathy}  
It measures the average empathy score of a speaker \(s\)’s messages, calculated as:  
\[
\text{empathy}(s) = \frac{\sum_{\mathcal{M} \in \mathcal{M}_s} EP(\mathcal{M})}{|\mathcal{M}_s|},
\]  
where \(EP(\mathcal{M})\) is the empathy of message \(\mathcal{M}\).

\begin{figure*}[t!]
    \centering
    \renewcommand{\thesubfigure}{}
    \setlength{\tabcolsep}{0pt} 
    \begin{tabular}{@{}cccc@{}}
        \subfigure{\includegraphics[width=0.24\textwidth]{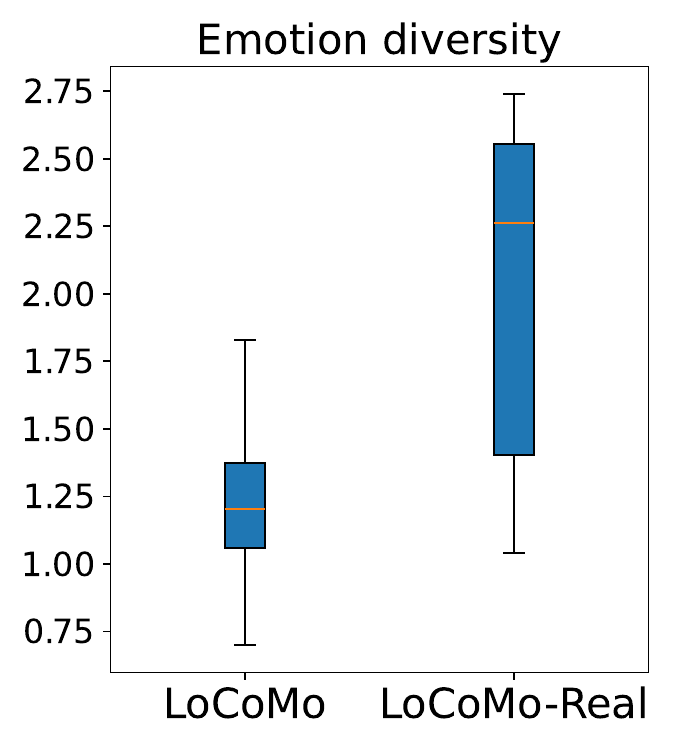}} &
        \subfigure{\includegraphics[width=0.24\textwidth]{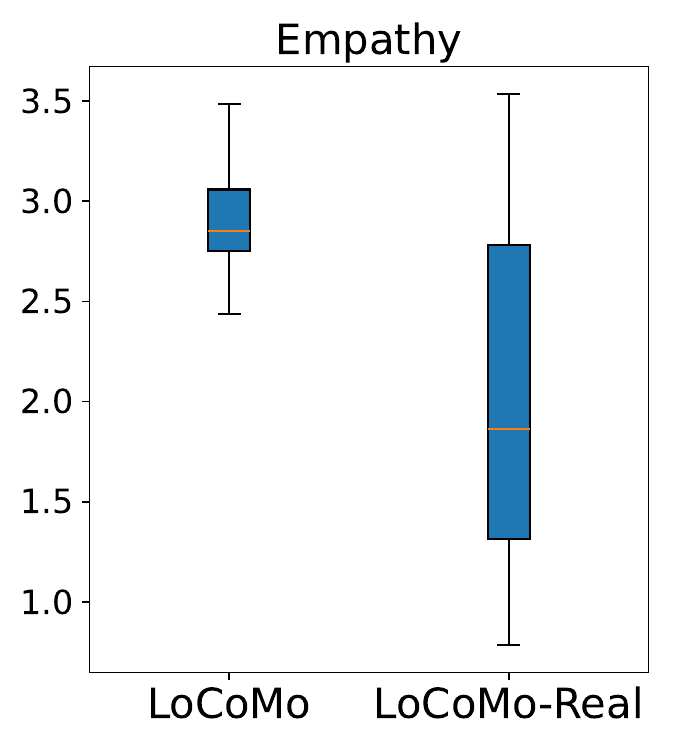}} &
        \subfigure{\includegraphics[width=0.24\textwidth]{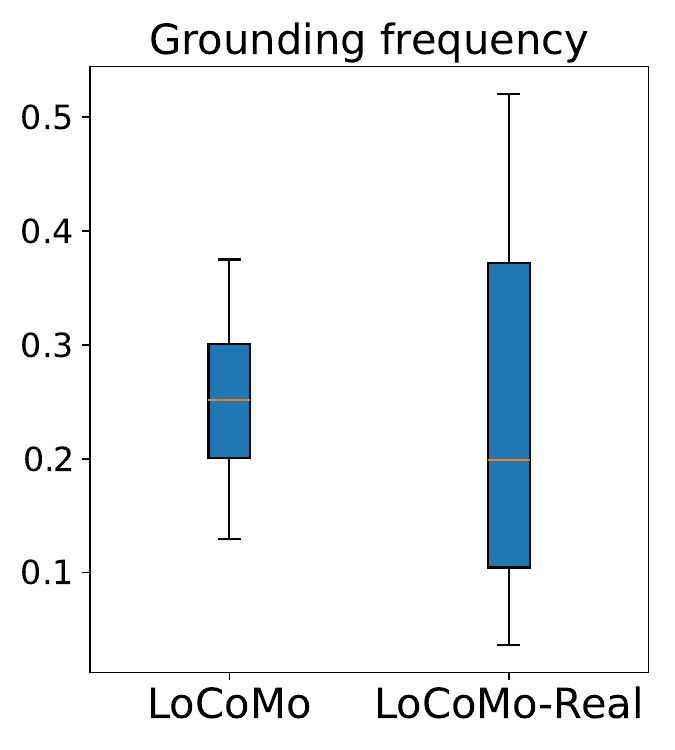}} &
        \subfigure{\includegraphics[width=0.24\textwidth]{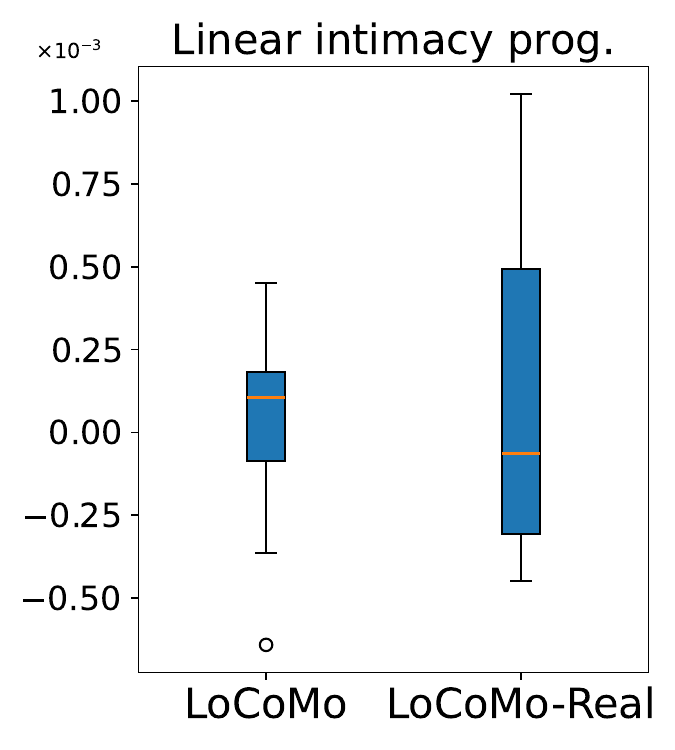}} \\
    \end{tabular}
    \vspace{-0.3cm}
    \caption{\textbf{Speaker-Level EI Comparison between \dataset{} and \textsc{LoCoMo}~\cite{maharana-etal-2024-evaluating}.} 
    Each bar represents the distribution of EI attributes for individual speakers in the conversation. 
    For a comprehensive evaluation, refer to Appendix~\ref{appendix:speaker-ei}.}
    \label{fig:speaker-level-ei}
    \vspace{-0.3cm}
\end{figure*}

\section{Data Statistics and Analysis}
\label{sec:analysis}
This section analyzes the collected data, starting with general observations and statistics (\secref{ssec:analysis-data}).  
Next, we compare our dataset with LLM-generated dialogues, focusing on speaker-level emotional intelligence (EI) metrics (\secref{ssec:analysis-speaker-level-ei}).  
Finally, we examine persona consistency in real-world conversations, leveraging the fact that each speaker participates in at least two conversations (\secref{ssec:analysis-persona}).  

\subsection{Data Statistics \& General Observations}  
\label{ssec:analysis-data}  
The dataset comprises 10 participant pairs engaging in 16–21 days of conversation, with daily word counts ranging from 691 to 861.
Participants share 23–46 images per pair and revisit topics 6–8 times, showcasing sustained and rich interactions.
Detailed statistics are provided in Appendix~\ref{appendix:chat-statistics}.
Three key characteristics distinguish real-world human dialogues from LLM-generated dialogues:
First, Human dialogues are inherently noisy, featuring typos, abbreviations, acronyms, and slang (\textit{e.g.}, ``imo'' for ``in my opinion'' and ``dunno'' for ``don’t know''), reflecting the informal and natural flow of communication.
Second, human conversations exhibit noticeable gaps between messages unlike LLM-generated dialogues, reflecting asynchronous and flexible communication.
On average, participants take 20.98 minutes to respond, with a median gap of 2.22 minutes.
Some gaps extend up to 27.90 hours, highlighting the non-linear nature of real-world interactions (Appendix~\ref{appendix:temporal-gap}).
Third, human dialogues often include varying lengths of consecutive messages (\textit{i.e.,} chat bubbles) by the same speaker, reflecting diverse communication patterns.
On average, participants send 2.31 consecutive messages per turn, with a median of 2.00 and a maximum of 68, showcasing occasional extended monologues (Appendix~\ref{appendix:chat-bubbles}).

\subsection{vs. LLM-generated Dialogues}  
\label{ssec:analysis-speaker-level-ei}  
We evaluate speaker-level EI (\secref{ssec:speaker-level-ei}) in real-world and LLM-generated dialogues using the \textsc{LoCoMo} dataset~\cite{maharana-etal-2024-evaluating}.
Speaker-level EI is derived from message-level EI: reflectiveness, grounding acts, and empathy are computed using \texttt{gpt-4o-mini}, while sentiment, emotion, and intimacy are classified using task-specific fine-tuned RoBERTa models trained on labeled Twitter data~\cite{antypas2023supertweeteval}.  
Each speaker in each conversation receives an independent EI score.  
Figure~\ref{fig:speaker-level-ei} illustrates key differences between human (\dataset{}) and LLM-generated (\textsc{LoCoMo}) dialogues by showing the distribution of speaker EI scores (full comparison in Appendix~\ref{appendix:speaker-ei}):
(1) Humans exhibit greater emotion and sentiment diversity, while LLMs remain constrained;
(2) LLMs display excessive empathy; and
(3) Humans show high variance in EI attributes, whereas LLMs are more uniform, aligning with prior research~\cite{lee2024language}.  

High emotional intelligence—characterized by strong grounding, emotional alignment, and intimacy progression—enhances engagement and relationship-building~\cite{altman1973social, hatfield1993emotional, derks2008role, huang2017doesn, niederhoffer2002linguistic}.  
However, since human EI naturally varies across individuals and interactions, enforcing uniformly high EI in LLMs does not make them more human-like.  
Instead, tailoring LLMs to specific individuals better reflects the diverse dynamics of real human interactions.

\subsection{Persona Consistency Analysis}  
\label{ssec:analysis-persona}  
\begin{figure}[t!]
    \centering
    \setlength{\tabcolsep}{0pt} 
    \includegraphics[width=\columnwidth]{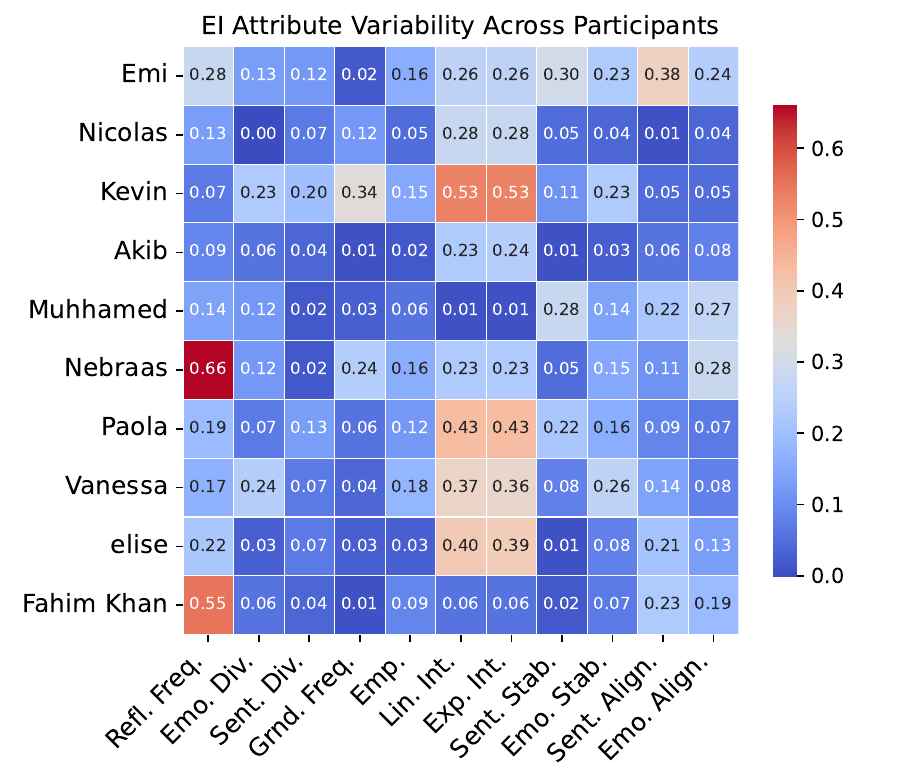}
    \vspace{-0.5cm}
    \caption{\textbf{Persona Consistency.} Each value represents the absolute difference in EI attributes for individual speakers across two conversations, capturing how their EI varies depending on their conversational partners.}
    \label{fig:persona-consistency}
    \vspace{-0.5cm}
\end{figure}

We examine how speakers adapt their emotional intelligence (EI) based on their conversational partners.  
Figure~\ref{fig:persona-consistency} shows absolute differences in EI attributes per participant, capturing persona variability across interactions:
(1) Some participants maintain a stable persona, while others dynamically adapt their EI; and
(2) Intimacy progression shows the most variation, suggesting that rapport-building is highly influenced by the specific conversational partner.
These findings indicate that a speaker's persona is flexible and adjusts according to social dynamics, reflecting adaptive emotional behavior.

\section{\dataset{} Benchmark}  
To test LLMs in long-term dialogues and advance human-like AI, we introduce: 
(1) \textbf{Persona simulation}, testing how well a model replicates an individual’s conversational style (\secref{ssec:persona-simulation}), and  
(2) \textbf{Memory probing}, assessing its ability to apply long-term context to answer probing questions (\secref{ssec:memory-probing}).

\subsection{Task 1: Persona Simulation}
\label{ssec:persona-simulation}

\begin{table*}[!t]
    \centering
    \resizebox{\textwidth}{!}{%
        \begin{tabular}{lcccccccc}
            \toprule
            & \multicolumn{2}{c}{\textbf{Content Similarity}} & \multicolumn{6}{c}{\textbf{Message-level EI (Emotional Intelligence)}} \\
            \cmidrule(lr){2-3} \cmidrule(lr){4-9}
            & Lexical $\uparrow$ & Semantic $\uparrow$ & Reflective $\uparrow$ & Grounding $\uparrow$ & Sentiment $\uparrow$ & Emotion $\uparrow$ & Intimacy $\downarrow$ & Empathy $\downarrow$ \\
            \midrule
            w/o fine-tune & \textbf{0.14 ± 0.04} & 0.76 ± 0.08  & 0.62 ± 0.13 & 0.40 ± 0.13 & 0.53 ± 0.22 & 0.43 ± 0.22 & \textbf{0.06 ± 0.01} & 1.80 ± 0.55 \\
            w/ fine-tune  & \textbf{0.14 ± 0.05} & \textbf{0.78 ± 0.04} & \textbf{0.77 ± 0.09} & \textbf{0.62 ± 0.08} & \textbf{0.59 ± 0.18} & \textbf{0.46 ± 0.21} & 0.07 ± 0.01 & \textbf{1.24 ± 0.12} \\
            \bottomrule
        \end{tabular}
    }
    \vspace{-0.2cm}
    \caption{\textbf{Average content similarity and message-level EI comparison in persona simulation} for all speakers, with and without fine-tuning. Full results are available in Appendix~\ref{appendix:persona-simulation}. Content similarity differences are not statistically significant, while message-level EI differences are significant (\( p < 0.02 \) for all attributes).}
    \label{tab:persona-simulation}
    \vspace{-0.3cm}
\end{table*}

\subsubsection{Experimental Setup.}
We simulate a speaker’s next message \(\hat{\mathcal{M}}_t\) based on prior conversation history \(\mathcal{H}_t = \{\mathcal{M}_1, ..., \mathcal{M}_{t-1}\}\) and compare it to the ground truth \(\mathcal{M}_t\).
Each speaker participates in two conversations (\(\mathcal{C}_a\) and \(\mathcal{C}_b\)), allowing us to evaluate two baselines:  
(1) \textbf{w/o fine-tuning}: A generic LLM tested on \(\mathcal{C}_b\); and
(2) \textbf{w/ fine-tuning}: An LLM fine-tuned on \(\mathcal{C}_a\) and tested on \(\mathcal{C}_b\).
We choose the conversation with lower overall EI as \(\mathcal{C}_b\).
During both training and testing, we use the prompt ``\textit{You are \{speaker name\}. Continue the conversation.}'' along with the previous conversation history.
The speaker’s original message serves as the ground truth for this prompt (See Appendix~\ref{ssec:appendix-persona-simulation} for prompt details).
To measure how effectively the model simulates the speaker, we compare message-level EI attributes (\secref{ssec:message-level-ei}) of predicted and ground truth messages using  
(1) accuracy for categorical and boolean attributes (\textit{i.e.,} reflectiveness, grounding act, sentiment, and emotion); and  
(2) absolute difference for continuous attributes (\textit{i.e.,} intimacy, empathy).
Lower absolute difference and higher accuracy indicate stronger simulation accuracy.  
Additionally, we compare lexical and semantic similarity between \(\hat{\mathcal{M}}_t\) and \(\mathcal{M}_t\), using ROUGE~\cite{lin-2004-rouge} and BERTScore~\cite{Zhang*2020BERTScore:}.

\subsubsection{Experimental Results.}
We explore (1) the impact of conversational context, (2) its role during fine-tuning; and (3) the impact of fine-tuning on a specific speaker.

\paragraph{Impact of conversational context.}  
We analyze how performance changes as the amount of provided context varies to assess the impact of conversational context.
Figure~\ref{fig:simulation-by-context} in Appendix~\ref{appendix:persona-simulation-context} shows that increasing conversation history does not improve message-level EI or exhibit a clear pattern.  
These results suggest that LLMs struggle to capture and replicate a speaker’s style using context alone.

\paragraph{Impact of fine-tuning.} 
We explore whether fine-tuning improves an LLM’s ability to simulate a speaker’s responses given conversational context.
A potential approach is to train on all \( C_a \) conversations and evaluate on \( C_b \).  
However, this risks data leakage, as a speaker in \( C_a \) may appear in \( C_b \).  
To ensure fairness, we train and test each speaker separately.  
First, we train models using different amounts of conversation history as context and test them with the same number of sessions used in training.
However, we observe no clear pattern indicating that increasing the amount of conversational context improves performance (See Appendix~\ref{appendix:persona-simulation-context}).
The results suggest that performance saturates after three sessions, implying that additional context beyond this point does not provide further benefits.

However, when comparing a fine-tuned model trained with three sessions of context to a non-fine-tuned version, we find that the model does learn the speaker’s unique style.
Table~\ref{tab:persona-simulation} presents the average performance across all speakers, with full results provided in Appendix~\ref{appendix:persona-simulation}. 
The findings highlight three key insights:
(1) Fine-tuning effectively captures a speaker’s style, including emotion, sentiment, reflectiveness, grounding, and empathy; 
(2) While stylistic adaptation improves, content similarity (lexical and semantic) remains largely unaffected;  
(3) Fine-tuning does not enhance intimacy, as intimacy emerges from mutual interaction rather than a single speaker’s style.

Overall, the results suggest that while the model does not learn how to use conversational context to improve its simulation, it can learn the stylistic patterns of a speaker when trained exclusively on their messages.

\subsection{Task 2: Memory Probing}  
\label{ssec:memory-probing}  

\subsubsection{Experimental Setup}  
We evaluate model performance on memory probing questions using long dialogues \( \mathcal{C} \).  
We compare two baselines:  
(1) \( \mathcal{C} \): The full conversation is provided as input.  
(2) \( \mathcal{E} \): Only human-annotated key events are provided, simulating a scenario where important details are stored as statements, a common feature in recent closed LLM APIs\footnote{\href{https://help.openai.com/en/articles/8590148-memory-faq}{OpenAI Memory FAQ}, \href{https://openai.com/index/memory-and-new-controls-for-chatgpt}{OpenAI Memory and Controls}}.
We report partial exact match F1 score following prior works~\cite{kwiatkowski-etal-2019-natural, maharana-etal-2024-evaluating} and LLM-based accuracy, where \texttt{gpt-4o-mini} determines whether the predicted answer matches the ground truth.
See Appendix~\ref{ssec:appendix-memory-probing} for prediction and evaluation prompt details.

\subsubsection{Experimental Results.}  
Table~\ref{tab:memory-probing} presents memory probing results comparing full conversation context (\(\mathcal{C}\)) and event-based memory (\(\mathcal{E}\)).  

\paragraph{Overall performance.}  
LLMs struggle with long-term memory even with full conversation (\(\mathcal{C}\)). 
While full context improves performance, models achieve only moderate accuracy, highlighting their limitations in applying long-term information. 

\paragraph{Event-based memory.}  
Event-based memory (\(\mathcal{E}\)) enhances efficiency but severely weakens memory probing performance, especially in multi-hop reasoning by causing a 41-46\% performance drop.  
By condensing conversations into key points, it often removes intermediary steps crucial for linking past and present context.  
Unlike temporal or commonsense reasoning, multi-hop tasks require reconstructing implicit connections, which event-based memory fails to retain.  
This tradeoff mirrors real-world AI memory challenges, such as memory system in LLM APIs, which stores key facts but may lose essential context for multi-turn reasoning.

\begin{table}[t!]
    \centering
    \resizebox{\columnwidth}{!}{%
        \begin{tabular}{lccccccc}
            \toprule
            \textbf{Model} & \textbf{Input} & \multicolumn{3}{c}{\textbf{Partial Match F1}} & \multicolumn{3}{c}{\textbf{LLM Accuracy}} \\
            \cmidrule(lr){3-5} \cmidrule(lr){6-8}
            &  & M & T & C & M & T & C \\
            \midrule
            gpt-4o-mini & $\mathcal{C}$ & 0.301 &	0.355 &	0.190 &	0.528	& 0.549 &	0.599  \\
            & $\mathcal{E}$ & 0.177 &	0.266 &	0.142 &	0.253 &	0.424 &	0.400 \\
            \midrule
            gpt-4o & $\mathcal{C}$ & 0.348 &	0.437 &	0.221 &	0.519 &	0.737 &	0.621 \\
            & $\mathcal{E}$ & 0.186 &	0.271 &	0.138 &	0.266 &	0.501 &	0.348 \\
            \bottomrule
        \end{tabular}
    }

    \caption{\textbf{Partial match F1 \& LLM Accuracy} for memory probing QA task. (M=\textsc{Multi-hop}, T={Temporal}, C=\textsc{Commonsense})}
    \label{tab:memory-probing}
\end{table}

\section{Conclusion}  
In this work, we introduce \dataset{}, a 21-day corpus of authentic messaging app dialogues—the longest available dataset where the same two individuals engage in sustained conversations. 
Through a detailed analysis, we highlight key differences between human conversations and LLM-simulated dialogues, revealing that LLMs exhibit limited emotion diversity, excessive empathy, and reduced variability in EI expression compared to real interactions.
Using this data, we present two benchmarks:  
(1) \textbf{Persona simulation}, evaluating LLMs' ability to replicate an individual’s conversational style and showing that simple fine-tuning improves simulation accuracy.  
(2) \textbf{Memory probing}, assessing LLMs' limitations in applying real-world long-term context and highlighting challenges in existing memory systems.  
Our findings underscore the complexity of modeling real-world persona dynamics and memory, emphasizing the need for further research.  
We hope \dataset{} inspires advancements in socially intelligent AI, personalized dialogue systems, and adaptive memory strategies, driving more human-like interaction modeling.  
\section{Limitations}  
Our study has several limitations:  

\paragraph{Demographic constraints.}  
Participants are English speakers from the US, aged 18-25, limiting the dataset’s representativeness across diverse demographics.  
Additionally, we do not analyze how conversational dynamics vary based on factors such as gender, profession, geographic location, or age differences.  
Future studies should include a more diverse demographic range to better understand how factors like age, gender, and cultural background influence dialogue patterns.

\paragraph{Chit-chat focus.}  
Our dataset primarily consists of casual, open-domain conversations rather than goal-oriented interactions.  
This limits its applicability to domains where social intelligence involves structured reasoning, such as negotiations or counseling~\cite{zhou2023sotopia, wu2024longmemeval, liu-etal-2024-interintent}.  
Future work should extend EI modeling to structured dialogues where speakers need to achieve their social goals.  

\paragraph{Emotional intelligence (EI) measurement.}  
Our EI evaluation focuses on surface-level indicators, such as sentiment, reflectiveness, and empathy.  
However, these metrics may not fully capture deeper cognitive, cultural, and situational influences on emotional intelligence.  
Additionally, EI attributes are inherently subjective and challenging to quantify.  
To ensure reliability, we ground our evaluations in established literature.  

\paragraph{Lack of multi-modal analysis.}  
While our dataset includes images, our experiments focus solely on text-based interactions.  
We do not analyze how visual elements contribute to emotional alignment, persona perception, or memory retention.  
Future research could integrate multi-modal fusion techniques to enhance emotional intelligence modeling.

\section{Ethical Considerations}
We took several steps to ensure that our data collection was ethical and legal. We set the hourly rate of compensation for workers at \$20.
To ensure the safety and well-being of our workers, we maintained open communication channels, allowing them to voice any question, concerns, or feedback about the data annotation.
This also helped to improve the quality of the collected data as we promptly addressed issues reported by workers throughout the process.

\subsubsection*{Acknowledgments}
Snap Inc. provided the majority of the funding for this work, with additional partial support from the Defense Advanced Research Projects Agency (DARPA) under award HR00112220046.
We also used Sahara AI’s data service platform for dataset construction.

We would like to thank Mohit Bansal (UNC), Yuwei Fang (Snap Inc.), Sergey Tulyakov (Snap Inc.) for their valuable discussions and contributions to this work.

\bibliography{custom}
\clearpage

\appendix
\section{Dataset Details}

\subsection{Participants}
\label{appendix:chat-participant}
All participants are native English speakers from the US, aged between 18 and 25, with a roughly equal distribution of genders. Each participant provided a signed release form before the study began. Participant details are shown in Table~\ref{tab:chat-participant}.

\begin{table}[h] 
\centering
\resizebox{0.9\columnwidth}{!}{%
\begin{tabular}{lllc}
\toprule
\textbf{Name} & \textbf{Job}           & \textbf{City}     & \textbf{Age} \\
\midrule
Emi           & College student                & New York          & 20           \\
Elise         & College student                & Houston           & 21           \\
Kevin         & College student                & Houston           & 18           \\
Paola         & College student                & Cambridge         & 21           \\
Nebraas       & Vet technician         & New York          & 24           \\
Nicolas       & College student                & New York          & 23           \\
Vanessa       & Vet technician         & New York          & 24           \\
Mohammed      & College student                & New York          & 23           \\
Syed          & College student                & New York          & 23           \\
Fahim         & College student                & New York          & 19           \\
\bottomrule
\end{tabular}%
}
\caption{\textbf{Participants Information}}
\label{tab:chat-participant}
\vspace{-0.3cm}
\end{table}

\subsection{Chat Statistics}
\label{appendix:chat-statistics}
Eight out of ten chat sessions have reached the target word count of 16,000. 
The data has been cleaned, with all incorrectly formatted links and images replaced. Images from external sources are attached in their original resolution (3024x4032). 
Participant-provided images from their camera rolls, which were unexpected, have been exported directly from WhatsApp without any modifications.
Statistics details are in presented in Table~\ref{tab:chat-statistics}.

\begin{table}[h!] 
\centering
\resizebox{\columnwidth}{!}{%
    \begin{tabular}{lccccc}
    \toprule
     \textbf{Chat} & \textbf{\# days} & \textbf{\# words} & \textbf{\# words / day} & \textbf{Images} & \textbf{Topics} \\ \midrule
    Emi + Elise         & 21 & 17341 & 826 & 35 & 6 \\
    Elise + Kevin       & 21 & 16040 & 764 & 46 & 7 \\ 
    Kevin + Paola       & 16 & 11057 & 691 & 35 & 7 \\
    Paola + Emi         & 16 & 11511 & 719 & 37 & 6 \\ 
    Nicolas + Nebraas   & 21 & 16902 & 805 & 37 & 7 \\ 
    Vanessa + Nicolas   & 21 & 18005 & 857 & 39 & 8 \\ 
    Vanessa + Nebraas   & 21 & 16343 & 778 & 43 & 8 \\ 
    Akib + Muhammed     & 21 & 17191 & 819 & 25 & 6 \\ 
    Fahim + Akib        & 21 & 18089 & 861 & 27 & 7 \\ 
    Fahim + Muhammed    & 21 & 16951 & 807 & 23 & 8 \\ 
    \bottomrule
    \end{tabular}
}
\captionsetup{justification=centering} 
\caption{\textbf{Chat Statistics}}
\label{tab:chat-statistics}
\vspace{-0.3cm}
\end{table}

\subsection{QA \& Event Statistics}
\label{appendix:qa-statistics}
We recruited a separate group of annotators to create memory-probing QA pairs and identify key events for each session $S_i$.
In total, 728 questions were annotated with answers, including 302 multi-hop, 321 temporal reasoning, and 111 commonsense questions.
Additionally, 600 events were annotated for each speaker in each session (see Table~\ref{tab:qa-statistics}).

\begin{table}[h!] 
\centering
\resizebox{\columnwidth}{!}{%
    \begin{tabular}{lccccc}
    \toprule
    \textbf{Chat} & \textbf{\textsc{Multi-Hop}} & \textbf{\textsc{Temporal}} & \textbf{\textsc{Commonsense}} & \textbf{\textsc{Events}} \\ \midrule
    Emi + Elise         & 30 & 30 & 10 & 44 \\
    Elise + Kevin       & 30 & 31 & 12 & 41 \\
    Kevin + Paola       & 30 & 30 & 11 & 51 \\
    Paola + Emi         & 30 & 30 & 10 & 61 \\
    Nicolas + Nebraas   & 31 & 31 & 14 & 61 \\
    Vanessa + Nicolas   & 30 & 30 & 12 & 84 \\
    Vanessa + Nebraas   & 30 & 30 & 12 & 90 \\
    Akib + Muhammed     & 30 & 31 & 10 & 83 \\
    Fahim + Akib        & 30 & 31 & 9 & 29 \\
    Fahim + Muhammed    & 30 & 45 & 10 & 56\\
    \bottomrule
    \end{tabular}
}
\captionsetup{justification=centering} 
\caption{\textbf{QA \& Event Statistics}}
\vspace{-0.3cm}
\label{tab:qa-statistics}
\end{table}

\subsection{Temporal gap between speaker}
\label{appendix:temporal-gap}
\begin{figure}[h]
    \centering
    \includegraphics[width=\columnwidth]{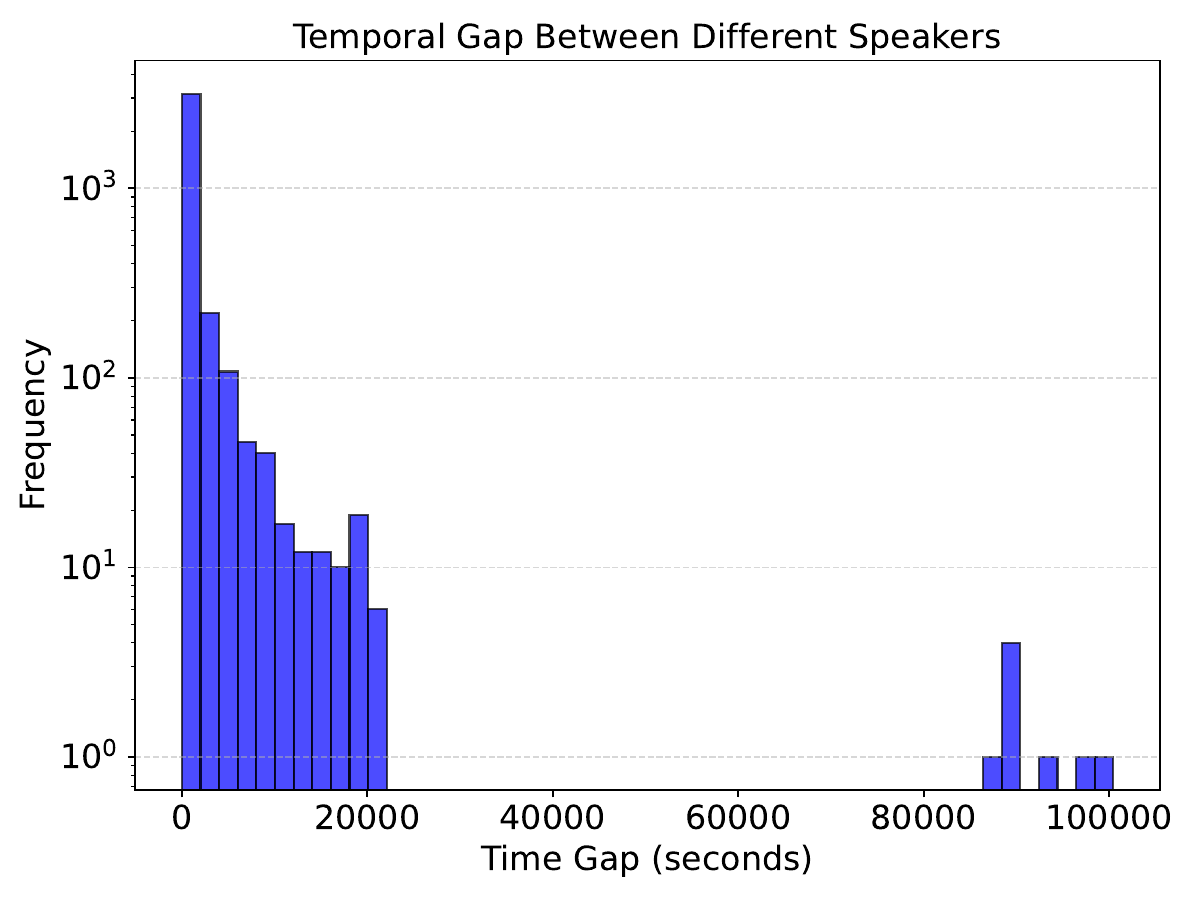}
\caption{\textbf{Temporal gap distribution} between messages from two speakers}
\vspace{-0.3cm}
\label{fig:temporal_gap_distribution}

\end{figure}

Figure~\ref{fig:temporal_gap_distribution} shows the distribution of temporal gaps between messages from two speakers.

\subsection{Lengths of consecutive messages}
\label{appendix:chat-bubbles}
\begin{figure}[h]
    \centering
    \includegraphics[width=\columnwidth]{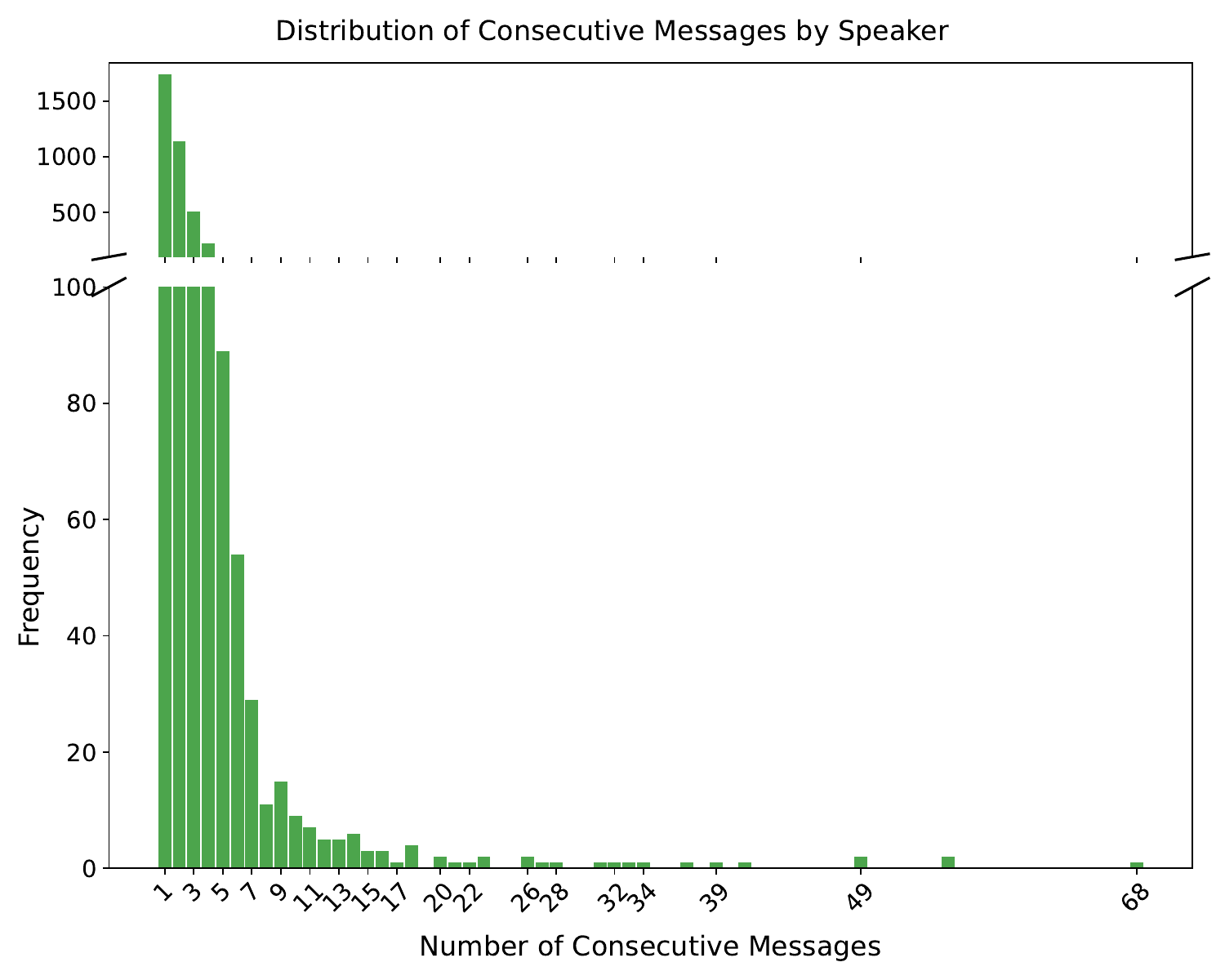}
    \caption{\textbf{Frequency distribution} of consecutive messages by a single speaker.}    \label{fig:chat_bubble_distribution}
\end{figure}

Figure~\ref{fig:chat_bubble_distribution} illustrates the distribution of consecutive messages per speaker, highlighting the frequency of different sequence lengths.

\onecolumn
\section{Annotation Guidelines}
\enlargethispage{1\baselineskip}

\subsection{Chat Collection Guidelines}
\label{appendix:chat-guidelines}
\nopagebreak

\begin{tcolorbox}[title=Chat Guidelines for Participants, myboxstyle, breakable, before skip=0pt]
\begin{description}[align=left, labelwidth=5em, labelindent=0em, itemsep=1em]
    \item[\textbf{First-Time Interaction.}]
    You are interacting with the other speaker for the first time.

    \item[\textbf{Chit-chat.}]
    Engage in chit-chat that can include real events from your own life (\textit{e.g.,} taking a nap or cooking something).
    \textit{The content can be fictional.}

    \item[\textbf{Personal Information.}]
    Make the conversation personal from time to time by discussing topics like family, friends, likes, dislikes, and aspirations.
    \textit{The content can be fictional.}

    \item[\textbf{Time References.}]
    Include references to time (\textit{e.g.,} ‘last Friday’, ‘next month’, ‘when I was ten years old’) and specific places or locations.
    Consider the current time during the conversation; for example, if it’s after lunch, ask what the other participant had for lunch, or greet them with “Good Morning” if chatting in the morning.

    \item[\textbf{Dialog Style.}]
    Keep the dialogue style casual, as if talking to a friend.

    \item[\textbf{Daily Events and News.}]
    Discuss events from your life, news, social media highlights, or pop culture events (\textit{e.g.,} movies, concerts).
    Feel free to share opinions in a friendly, engaging way that may interest the other participant.

    \item[\textbf{Images.}]
    Share images at appropriate points in the conversation. Examples include:
    \begin{itemize}
        \item things you own (clothes, food, vehicles, etc.),
        \item things you've seen or want to see (\textit{e.g.,} food, people, events),
        \item things you like (\textit{e.g.,} a piece of clothing, a cute animal),
        \item things you need help with (\textit{e.g.,} how to fix something or navigate a relationship),
        \item images representing old memories.
    \end{itemize}

    \item[\textbf{Image Source.}]
    Do not share images from your camera. Instead, search for images on the internet using Google, save the image URL, and include it in square brackets within the conversation.
    \begin{quote}
        \texttt{[URL of the image]}
    \end{quote}
    Images should be licensed under Creative Commons to allow free use by others. In Google Search, such images can be found by selecting the Creative Commons license under Tools → Usage Rights. Using the dimension constraint \texttt{imagesize:3024x4032} in the query yields less generic images that can foster engaging conversations.

    \item[\textbf{Images Containing Faces.}]
    Avoid sharing images where a person's face is clearly visible and implies that it is you. However, you may share images where the face is obscured or clearly not you. Ensure any appearance remains approximately consistent throughout sessions.

    \item[\textbf{Image Context.}]
    Avoid sharing images that don’t add information to the conversation. For example, avoid generic images that lack context or relevance, but do share images that add context, like one showing a user kayaking.

    \item[\textbf{Pronoun Use.}]
    Refer to previously mentioned subjects in the conversation with pronouns, allowing the other participant to easily infer the reference.
\end{description}
\end{tcolorbox}

\subsection{Question \& Event Annotation Guidelines}
\label{appendix:qa-guidelines}

\begin{tcolorbox}[title=Question \& Event Annotation Guidelines, myboxstyle, breakable]
\begin{description}[align=left, labelwidth=5em, labelindent=0em, itemsep=1em]

\item[\textbf{Introduction.}] These guidelines provide instructions for annotating a dataset consisting of long-term conversations (approximately 21 days) between two individuals. The annotation task is divided into two sub-tasks:
    \begin{itemize}
        \item \textbf{Events Annotation:} Annotators document significant life events for each speaker in each session.
        \item \textbf{Question and Answer Creation:} Annotators generate questions that can only be answered by reading through one or more sessions of conversations.
    \end{itemize}
    The dataset includes both textual and image-based conversations.

\item[\textbf{Data Format.}] The conversation dataset consists of multiple interaction sessions between two speakers. Each session is timestamped.
    \begin{itemize}
        \item Conversations are structured in a table format with dedicated columns for each speaker.
        \item Speaker names appear at the top of the table.
        \item If a speaker shares an image, it is included in the dataset.
    \end{itemize}

\item[\textbf{Sub-Task 1: Writing Speaker Events.}] Annotators document key life events for each speaker as the conversation progresses. These events may include:
    \begin{itemize}
        \item Past events, ongoing events, and planned future events.
        \item Small-scale events (e.g., attending a cooking class) and major events (e.g., moving to a new city).
    \end{itemize}
    \textbf{Annotation Format:}  
    \begin{itemize}
        \item Each event should be recorded in a separate line within the allotted cell.
    \end{itemize}

\item[\textbf{Sub-Task 2: Creating Questions and Answers.}] Annotators generate questions that can only be answered by reviewing the conversation dataset.

    \textbf{Question Categories and Examples:}
    \begin{itemize}
        \item \textbf{Questions that Require Aggregation of Information.} \newline
            \textbf{Question:} Which countries has Joseph travelled to? \newline
            \textbf{Answer:} France, Japan \newline
            \textbf{Evidence:} D4:1, D1:8 \newline
            \textbf{Category:} 1
        \item \textbf{Questions that Require Reasoning About Time.} \newline
            \textbf{Question:} When did Kate start skiing? \newline
            \textbf{Answer:} 2013 \newline
            \textbf{Evidence:} D1:49 \newline
            \textbf{Category:} 2
        \item \textbf{Questions that Require Commonsense Reasoning or World Knowledge.} \newline
            \textbf{Question:} What kind of jobs might Joseph consider based on his recent ventures? \newline
            \textbf{Answer:} Jobs that combine software engineering and management, e.g., software product manager \newline
            \textbf{Evidence:} D2:9, D3:1 \newline
            \textbf{Category:} 3
    \end{itemize}

    \textbf{How to Annotate Questions:}  
    Each question entry includes the following fields:
    \begin{itemize}
        \item \textbf{Question:} The formulated question.
        \item \textbf{Answer:} The exact response extracted from the conversation.
        \item \textbf{Evidence:} Unique identifiers of dialogues used to derive the answer.
        \item \textbf{Category:} A label (1-3) indicating the question type.
    \end{itemize}
    Annotators should write at least \textbf{2 and up to 10 questions} from any of the categories at the end of each session.

\end{description}
\end{tcolorbox}

\section{Message-level EI Attributes}
In this section, we provide prompt details for reflectiveness, grounding act, and empathy.
Also, we present model details for emotion, sentiment, and intimacy.

\subsection{Prompt for `Reflectiveness'}
\label{appendix:evaluation-reflective}

\begin{tcolorbox}[title=Reflectiveness Classification, myboxstyle, breakable]
You are an evaluator trained to determine if a speaker’s language is reflective, indicating self-awareness. Reflective language is characterized by self-observation, perspective-taking, and intentionality. This means that the speaker is not only aware of their thoughts, feelings, or actions but also able to express this awareness clearly. \\

A reflective response often includes one or more of the following traits:

\begin{itemize}
    \item \textbf{Self-observation:} The speaker describes their own emotional or cognitive state (e.g., “I feel uncertain about…” or “I’m aware that…”).
    \item \textbf{Perspective-taking:} The speaker shows an understanding of how their actions or emotions affect others or acknowledges another person’s perspective on the situation (e.g., “I understand that my response may seem…”).
    \item \textbf{Intentionality:} The speaker explains the reasoning behind their behavior or decisions, revealing their underlying motivations or goals (e.g., “I decided to respond this way because…”).
\end{itemize}

\textbf{Example Statements}
\begin{itemize} 
    \item "I realize I tend to get defensive when I receive feedback, and I think it’s because I want to do well."  \\
    \textbf{Reflective or Not Reflective:} Reflective  \\
    \textbf{Reason:} This statement shows self-observation (“I realize I tend to get defensive”) and insight into motivation (“because I want to do well”).

    \item "I did what I thought was best for the project."  \\
    \textbf{Reflective or Not Reflective:} Not Reflective  \\
    \textbf{Reason:} While the speaker describes their decision, they don’t analyze or acknowledge the emotions or motivations behind their choice or consider its impact on others.
\end{itemize}

Given this dialogue context:  
\texttt{{\{dialogue\_history\_within\_session\}}}

Determine whether the \texttt{{speaker}}'s last message (\texttt{\{turn}\}) is reflective or not.  \\
Reflective language includes phrases like 'I feel...', 'I think...', or similar reflective expressions.  \\
Respond only with \texttt{'True'} for reflective or \texttt{'False'} for not reflective.
\end{tcolorbox}

\subsection{Prompt for `Grounding Act'}
\label{appendix:evaluation-grounding}

\begin{tcolorbox}[title=Grounding Act Classification, myboxstyle, breakable]
You are an evaluator trained to determine if a speaker’s language demonstrates grounding, which reflects active engagement and a commitment to mutual understanding in conversation. Grounding acts are characterized by clarifying questions, follow-up inquiries, or statements that seek to confirm, clarify, or expand on shared information. These acts are essential for building common ground, ensuring that both participants have a clear understanding, and preventing misunderstandings. \\

A grounding response often includes one or more of the following traits: \\

\textbf{Clarifying questions:} The speaker asks questions that seek clarification or further information about the other person’s statements (e.g., “Could you explain that further?” or “What did you mean by...?”). \\

\textbf{Follow-up inquiries:} The speaker shows interest in exploring a point raised by the other person, prompting them to elaborate or continue sharing (e.g., “How did that make you feel?” or “Can you tell me more about...?”). \\

\textbf{Confirmation checks:} The speaker seeks to confirm their understanding of what the other person said (e.g., “So, you mean that...?” or “Are you saying that...?”). \\

\section*{Example Statements}

\textbf{"Can you tell me more about what happened at the event?"} \\
Grounding or Not Grounding: \textbf{Grounding} \\
Reason: This is a follow-up question that prompts the other person to provide more information, demonstrating interest and a desire to deepen mutual understanding. \\

\textbf{"I completely understand your point."} \\
Grounding or Not Grounding: \textbf{Not Grounding} \\
Reason: Although this statement indicates agreement, it does not actively seek further information or clarification and does not encourage continued dialogue.\\

\textbf{"So, you’re saying that this new policy will impact the timeline?"} \\
Grounding or Not Grounding: \textbf{Grounding} \\
Reason: This is a confirmation check, as the speaker seeks to ensure their understanding of the other person’s statement.\\

\textbf{"It sounds like you’ve already made your decision."} \\
Grounding or Not Grounding: \textbf{Not Grounding} \\
Reason: This statement reflects an observation rather than a clarifying or follow-up question, so it does not serve as a grounding act.
\end{tcolorbox}

\subsection{Prompt for `\texttt{Empathy}'}
\label{appendix:evaluation-empathy}

\begin{tcolorbox}[title=Empathy Assessment, myboxstyle, breakable]
You are an evaluator assessing the level of empathy conveyed in a response, based on three core components: \textbf{Emotional Reaction}, \textbf{Interpretation}, and \textbf{Exploration}. For each component, provide a score from 0–2, where 0 indicates no presence, 1 indicates partial presence, and 2 indicates explicit presence. Sum the scores from each component to obtain an overall empathy score.

\section*{Component 1: Emotional Reaction}
Does the response express or allude to warmth, compassion, concern, or similar feelings of the responder towards the seeker?
\begin{itemize}
    \item \textbf{0}: No.
    \item \textbf{1}: Yes, the response alludes to these feelings but the feelings are not explicitly expressed.
    \item \textbf{2}: Yes, the response has an explicit mention.
\end{itemize}

\section*{Component 2: Interpretation}
Does the response communicate an understanding of the seeker’s experiences and feelings? In what manner?
\begin{itemize}
    \item \textbf{0}: No.
    \item \textbf{1}: Yes, the response communicates an understanding of the seeker’s experiences and/or feelings.
    \begin{itemize}
        \item The response contains conjectures or speculations about the seeker’s experiences and/or feelings.
        \item The responder reflects back on similar experiences of their own or others.
        \item The responder describes similar experiences of their own or others.
        \item The response paraphrases the seeker’s experiences and/or feelings.
    \end{itemize}
    \item \textbf{2}: The response provides a deep, explicit understanding and validation of the seeker’s feelings or experiences, potentially using multiple sub-categories.
\end{itemize}

\section*{Component 3: Exploration}
Does the response make an attempt to explore the seeker’s experiences and feelings?
\begin{itemize}
    \item \textbf{0}: No.
    \item \textbf{1}: Yes, the exploration is present but remains generic.
    \item \textbf{2}: Yes, the exploration is present and is specific, delving into the seeker’s particular feelings or experiences.
\end{itemize}

\section*{Output Format}
Return output in the following JSON format:
\begin{verbatim}
{
    "emotional_reaction": [0–2],
    "interpretation": [0–2],
    "exploration": [0–2]
}
\end{verbatim}

\end{tcolorbox}

\definecolor{systembg}{RGB}{230, 255, 230} 
\definecolor{userbg}{RGB}{255, 240, 230}   
\definecolor{assistantbg}{RGB}{230, 230, 255} 

\section{Persona Simulation \& Memory Probing}  
In this section, we detail the prompts used to evaluate LLM performance on persona simulation and memory probing tasks.  
We specify the assigned roles (\textit{i.e.,} system, user, assistant) for each prompt in our experiments.  

\subsection{Persona Simulation}
\label{ssec:appendix-persona-simulation}
The input consists of a system and user prompt, where the system defines the user’s persona and task, while the user prompt provides the dialogue history.  
The output serves as ground truth for training.

\subsubsection{Input}
\begin{tcolorbox}[colback=systembg, colframe=black, sharp corners]
\textbf{System:} \\
You are \texttt{\{opponent\_speaker\}}. Continue the conversation. \\
Output only the message, not the speaker name.
\end{tcolorbox}

\begin{tcolorbox}[colback=userbg, colframe=black, sharp corners]
\textbf{User:} \\
\texttt{\{previous conversation\}} \\
\texttt{\{speaker\}}
\end{tcolorbox}

\subsubsection{Output (Ground Truth)}
\begin{tcolorbox}[colback=assistantbg, colframe=black, sharp corners]
\textbf{Assistant:} \\
\texttt{\{speaker's original message\}}
\end{tcolorbox}

\subsection{Memory Probing}
\label{ssec:appendix-memory-probing}
For the memory probing task, we use only user role messages.  
All conversation history is formatted consistently, with sessions divided by date, as shown in the following example.  
If a message contains an image, we convert it into a caption using BLIP.  

\begin{tcolorbox}[colback=userbg, colframe=black, sharp corners]  
\textbf{User:} \\  
Below is a conversation between Alice and Bob, spanning multiple days. Each session starts with the corresponding date. \\  

\textbf{DATE:} February 10, 2025  

\textbf{CONVERSATION:}  \\  
Alice: "Hey, how have you been?"  \\  
Bob: "I'm good, just busy with work. How about you?"  \\  
Alice: "Same here, lots of meetings this week." \\  
Alice shared an image of "papers on the desk." \\  

Based on the provided context, generate a concise short-phrase answer for the following question.  
If the question pertains to a date, infer the approximate timeframe (e.g., "In the 1800s", "Before Jan 2021", etc.). \\  

\textbf{Question:} What has Alice been busy with?  

\textbf{Answer:}  
\end{tcolorbox}  

\section{Experimental Results}
\subsection{Full Performance of Speaker-level EI}
\label{appendix:speaker-ei}
\begin{figure*}[!h]
    \renewcommand{\thesubfigure}{}
    \captionsetup{justification=centering}
    \setlength{\tabcolsep}{0pt} 
    \begin{tabular}{@{}cccc@{}}
        \subfigure{\includegraphics[width=0.24\textwidth]{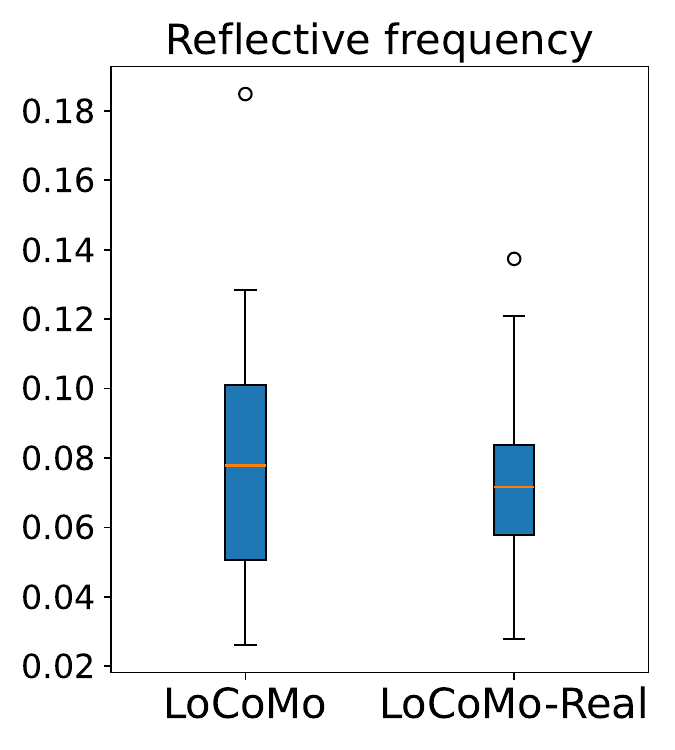}} &
        \subfigure{\includegraphics[width=0.24\textwidth]{figures/speaker_level_ei/emotion_diversity.pdf}} &
        \subfigure{\includegraphics[width=0.24\textwidth]{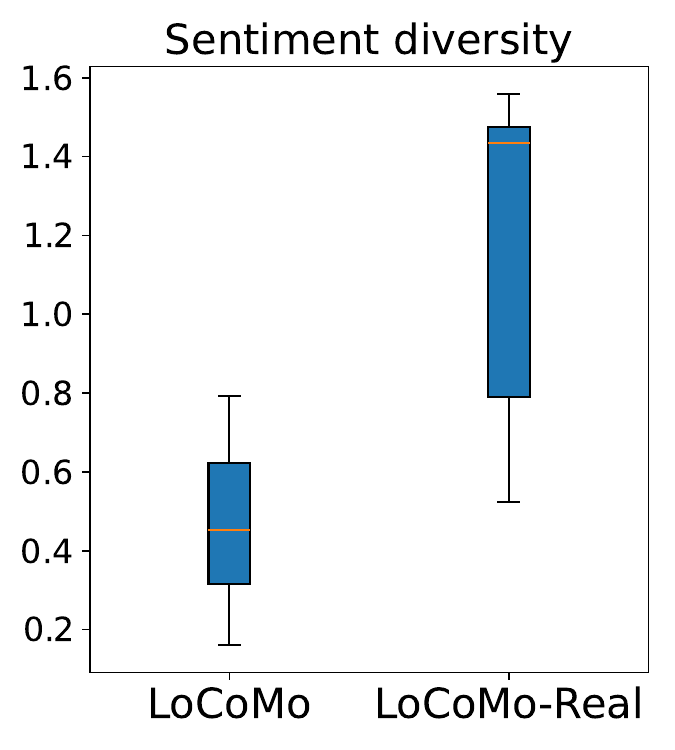}} &
        \subfigure{\includegraphics[width=0.24\textwidth]{figures/speaker_level_ei/grounding_frequency.pdf}} \\
        \subfigure{\includegraphics[width=0.24\textwidth]{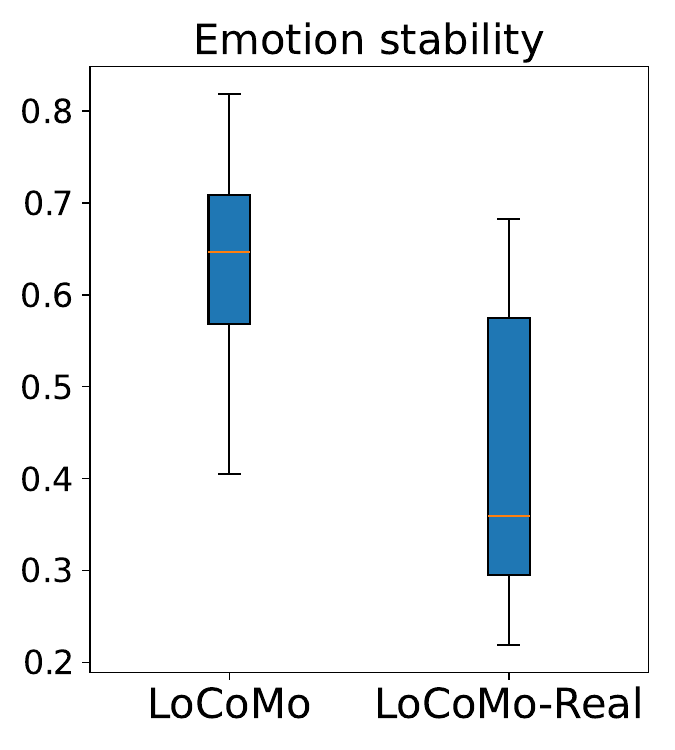}} &
        \subfigure{\includegraphics[width=0.24\textwidth]{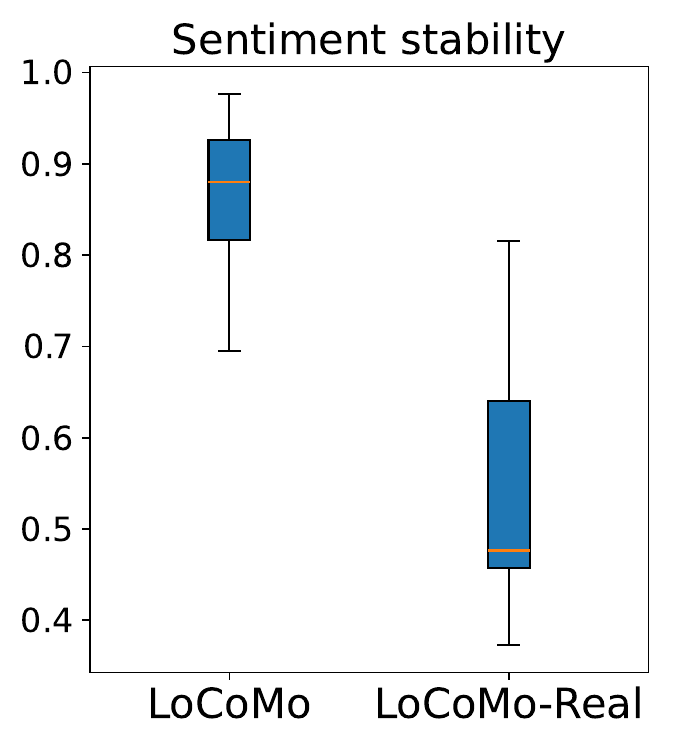}} &
        \subfigure{\includegraphics[width=0.24\textwidth]{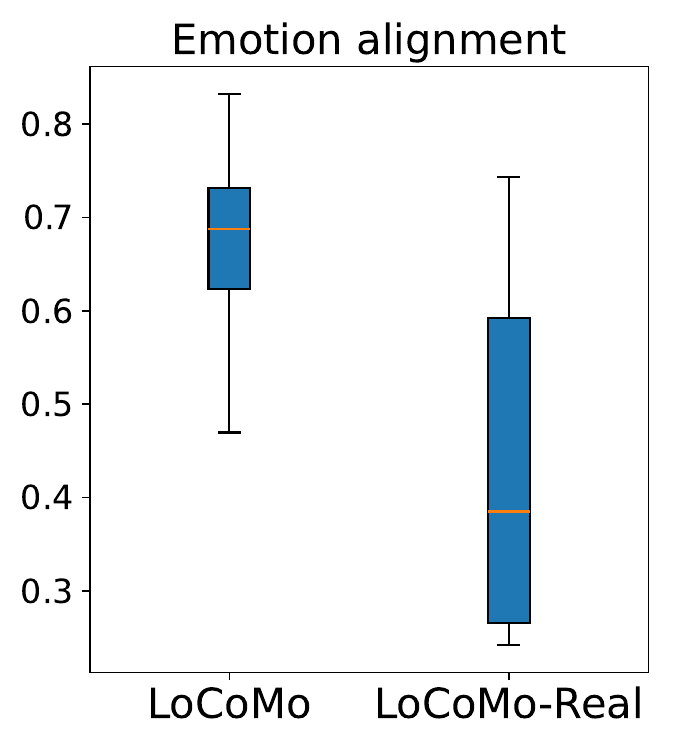}} &
        \subfigure{\includegraphics[width=0.24\textwidth]{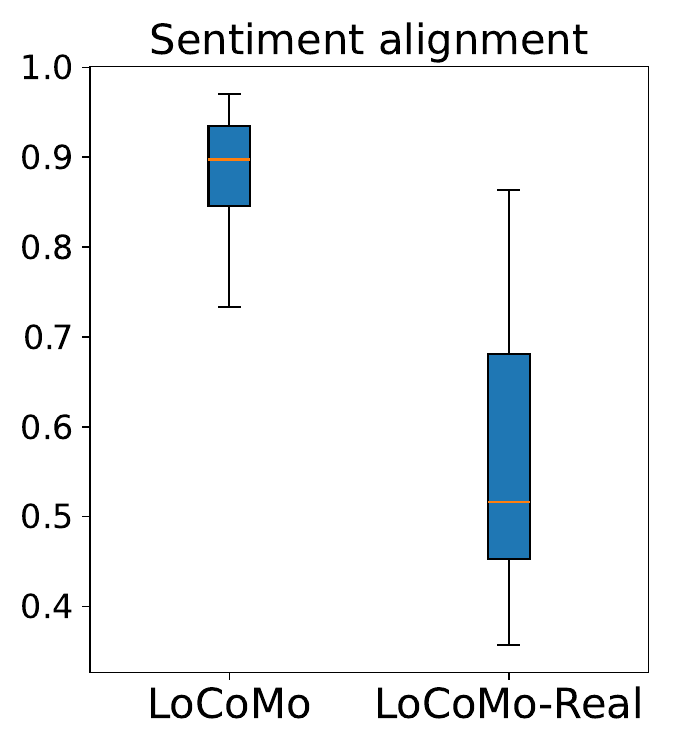}} \\
        \subfigure{\includegraphics[width=0.24\textwidth]{figures/speaker_level_ei/empathy.pdf}} &
        \subfigure{\includegraphics[width=0.24\textwidth]{figures/speaker_level_ei/linear_intimacy_progression.pdf}} &
        \subfigure{\includegraphics[width=0.24\textwidth]{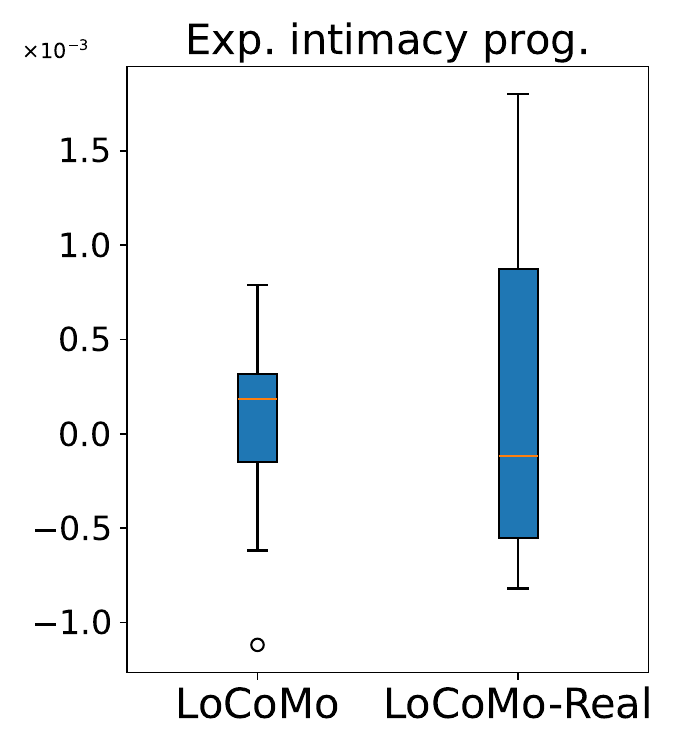}} &
    \end{tabular}
    \caption{\textbf{Full Speaker-Level EI Comparison between \dataset{} and \textsc{LoCoMo}~\cite{maharana-etal-2024-evaluating}.}  
    Each bar represents the distribution of EI attributes across individual speakers in the dataset.} 
    \label{fig:full-speaker-level-ei}
\end{figure*}

Figure~\ref{fig:full-speaker-level-ei} presents a full comparison of speaker-level EI between \dataset{} and \textsc{LoCoMo}~\cite{maharana-etal-2024-evaluating}, covering all 11 evaluated attributes.  
Table~\ref{tab:full-locomo-real-metrics} provides the detailed speaker-level EI metrics for \dataset{}, which serve as the basis for Figure~\ref{fig:full-speaker-level-ei}.

\begin{table*}[!h]
\centering
\resizebox{\textwidth}{!}{%
\begin{tabular}{llrrrrrrrrrrrrrrrrrrrrrr}
\toprule
              &               & \multicolumn{6}{c}{\textbf{Self-awareness}}                                                                                                                  & \multicolumn{2}{c}{\textbf{Motivation}}                  & \multicolumn{2}{c}{\textbf{Empathy}}                     & \multicolumn{4}{c}{\textbf{Social Skills} (1e-3)}                                                                  & \multicolumn{8}{c}{\textbf{Self-regulation}}                                                                                                                                                                   \\
\cmidrule(lr){3-8} \cmidrule(lr){9-10} \cmidrule(lr){11-12} \cmidrule(lr){13-16} \cmidrule(lr){17-24}
              &               & \multicolumn{2}{c}{Reflect. freq.}        & \multicolumn{2}{c}{Emo. div.}           & \multicolumn{2}{c}{Sent. div.}         & \multicolumn{2}{c}{Ground. freq.}         & \multicolumn{2}{c}{EPITOME}                     & \multicolumn{2}{c}{Int. prog. (lin)} & \multicolumn{2}{c}{Int. prog. (exp.)} & \multicolumn{2}{c}{Sent. stab.}         & \multicolumn{2}{c}{Emo. stab.}           & \multicolumn{2}{c}{Sent. align.}         & \multicolumn{2}{c}{Emo. align.}           \\
\cmidrule(lr){3-4} \cmidrule(lr){5-6} \cmidrule(lr){7-8} \cmidrule(lr){9-10} \cmidrule(lr){11-12} \cmidrule(lr){13-14} \cmidrule(lr){15-16} \cmidrule(lr){17-18} \cmidrule(lr){19-20} \cmidrule(lr){21-22} \cmidrule(lr){23-24}
\textbf{P1} & \textbf{P2} & \multicolumn{1}{c}{P1} & \multicolumn{1}{c}{P2} & \multicolumn{1}{c}{P1} & \multicolumn{1}{c}{P2} & \multicolumn{1}{c}{P1} & \multicolumn{1}{c}{P2} & \multicolumn{1}{c}{P1} & \multicolumn{1}{c}{P2} & \multicolumn{1}{c}{P1} & \multicolumn{1}{c}{P2} & \multicolumn{1}{c}{P1}  & \multicolumn{1}{c}{P2}  & \multicolumn{1}{c}{P1}     & \multicolumn{1}{c}{P2}    & \multicolumn{1}{c}{P1} & \multicolumn{1}{c}{P2} & \multicolumn{1}{c}{P1} & \multicolumn{1}{c}{P2} & \multicolumn{1}{c}{P1} & \multicolumn{1}{c}{P2} & \multicolumn{1}{c}{P1} & \multicolumn{1}{c}{P2} \\
\midrule

Fahim    & Muhhamed      & 0.14  & 0.11  & 2.21  & 2.41  & 1.43  & 1.53  & 0.11  & 0.04  & 1.54  & 1.50  & -0.2  & -0.4  & -0.4  & -0.7  & 0.46  & 0.50  & 0.38  & 0.32  & 0.47  & 0.55  & 0.36  & 0.40  \\
Emi           & Paola         & 0.08  & 0.06  & 1.11  & 1.15  & 0.67  & 0.52  & 0.43  & 0.44  & 3.23  & 3.53  & 0.6   & 0.5   & 1.1   & 0.9   & 0.75  & 0.82  & 0.68  & 0.66  & 0.86  & 0.78  & 0.74  & 0.67  \\
Akib          & Muhhamed      & 0.06  & 0.12  & 2.63  & 2.61  & 1.48  & 1.56  & 0.04  & 0.05  & 0.87  & 1.32  & -0.1  & -0.4  & -0.2  & -0.7  & 0.46  & 0.37  & 0.29  & 0.26  & 0.45  & 0.43  & 0.35  & 0.26  \\
Emi           & elise         & 0.05  & 0.07  & 1.33  & 1.53  & 0.80  & 1.03  & 0.44  & 0.28  & 2.80  & 2.29  & 1.0   & 1.0   & 1.8   & 1.7   & 0.61  & 0.58  & 0.58  & 0.51  & 0.67  & 0.57  & 0.62  & 0.55  \\
Nicolas       & Nebraas       & 0.04  & 0.10  & 2.74  & 2.62  & 1.44  & 1.44  & 0.15  & 0.27  & 0.92  & 1.86  & 0.1   & -0.4  & 0.1   & -0.6  & 0.44  & 0.46  & 0.24  & 0.25  & 0.45  & 0.49  & 0.24  & 0.24  \\
Vanessa       & Nebraas       & 0.09  & 0.03  & 2.08  & 2.41  & 1.46  & 1.41  & 0.19  & 0.15  & 2.35  & 1.43  & -0.3  & 0.0   & -0.6  & 0.0   & 0.47  & 0.48  & 0.42  & 0.32  & 0.46  & 0.54  & 0.29  & 0.38  \\
Kevin         & Paola         & 0.07  & 0.08  & 1.04  & 1.27  & 0.55  & 0.66  & 0.52  & 0.41  & 3.19  & 3.20  & -0.1  & -0.1  & -0.2  & -0.3  & 0.80  & 0.72  & 0.68  & 0.59  & 0.71  & 0.73  & 0.59  & 0.70  \\
Akib          & Fahim         & 0.07  & 0.08  & 2.53  & 2.32  & 1.44  & 1.47  & 0.05  & 0.11  & 0.92  & 1.29  & -0.4  & -0.3  & -0.8  & -0.5  & 0.45  & 0.47  & 0.31  & 0.34  & 0.48  & 0.36  & 0.39  & 0.26  \\
\bottomrule
\end{tabular}%
}
\caption{\textbf{Full Speaker-Level EI Datapoints of \dataset{}.}}  
\label{tab:full-locomo-real-metrics}
\end{table*}

\subsection{Full Performance of Persona Simulation}
\label{appendix:persona-simulation}
Table~\ref{tab:full-persona-simulation} provides the complete results for content similarity and message-level EI comparison in persona simulation across all speakers, with and without fine-tuning.  
Table~\ref{tab:persona-simulation} is derived by averaging these values.
Persona consistency in the table is measured as the absolute difference between a speaker’s average EI across two different conversations.  
This approximates the impact of persona consistency on simulation performance.  
Results indicate that simulation performance improves when the difference is small, suggesting that speakers with more consistent personas across conversations are easier to simulate.  

\begin{table*}[!h]
    \centering
    \resizebox{\textwidth}{!}{%
        \begin{tabular}{llccccccccccc}
            \toprule
            \multirow{2}{*}{\textbf{Speaker}} & \multirow{2}{*}{\parbox{2cm}{\textbf{Persona Consistency}}} & \multirow{2}{*}{\textbf{Train}} & \multirow{2}{*}{\textbf{Test}} & \multirow{2}{*}{\textbf{Finetune}} & \multicolumn{2}{c}{\textbf{Content Similarity}} & \multicolumn{6}{c}{\textbf{Style Similarity}} \\
            \cmidrule(lr){6-7} \cmidrule(lr){8-13}
            & & & & & Lexical $\uparrow$ & Semantic $\uparrow$ & Reflective $\uparrow$ & Grounding $\uparrow$ & Sentiment $\uparrow$ & Emotion $\uparrow$ & Intimacy $\downarrow$ & Empathy $\downarrow$ \\
            \midrule
            Emi & 0.21 & Emi-Paola & Emi-Elise & \xmark & 0.2 & \textbf{0.84} & 0.78 & 0.6 & \textbf{0.85} & 0.71 & \textbf{0.06} & 1.56 \\
            & & & & \cmark & 0.2 & \textbf{0.84} & \textbf{0.81} & \textbf{0.63} & \textbf{0.85} & \textbf{0.76} & 0.07 & \textbf{1.43} \\
            \midrule
            Nicolas & 0.09 & Nicolas-Nebraas & Vanessa-Nicolas & \xmark & 0.11 & \textbf{0.82} & 0.64 & 0.38 & 0.38 & 0.21 & \textbf{0.08} & 2.2 \\
            & & & & \cmark & 0.09 & 0.81 & \textbf{0.88} & \textbf{0.73} & \textbf{0.45} & \textbf{0.26} & 0.09 & \textbf{1.18} \\
            \midrule
            Kevin & 0.22 & Kevin-Paola & Kevin-Elise & \xmark & 0.19 & \textbf{0.8} & \textbf{0.76} & 0.52 & \textbf{0.86} & 0.68 & \textbf{0.05} & 1.32 \\
            & & & & \cmark & 0.19 & \textbf{0.8} & 0.72 & \textbf{0.53} & \textbf{0.86} & 0.64 & 0.06 & \textbf{1.27} \\
            \midrule
            Akib & 0.07 & Fahim-Akib & Akib-Muhhamed & \xmark & 0.09 & \textbf{0.72} & 0.53 & 0.25 & 0.25 & 0.15 & \textbf{0.05} & 2.33 \\
            & & & & \cmark & 0.09 & \textbf{0.72} & \textbf{0.61} & \textbf{0.72} & \textbf{0.43} & \textbf{0.24} & 0.07 & \textbf{1.38} \\
            \midrule
            Muhhamed & 0.11 & Fahim-Muhhamed & Akib-Muhhamed & \xmark & 0.1 & \textbf{0.73} & 0.46 & 0.26 & 0.39 & 0.29 & \textbf{0.06} & 2.36 \\
            & & & & \cmark & 0.1 & \textbf{0.73} & \textbf{0.72} & \textbf{0.68} & \textbf{0.46} & 0.29 & 0.07 & \textbf{1.22} \\
            \midrule
            Nebraas & 0.2 & Nicolas-Nebraas & Nebraas-Vanessa & \xmark & 0.15 & \textbf{0.81} & 0.74 & 0.35 & 0.5 & 0.37 & \textbf{0.07} & 2.0 \\
            & & & & \cmark & 0.15 & 0.8 & \textbf{0.87} & \textbf{0.66} & \textbf{0.53} & \textbf{0.42} & 0.09 & \textbf{1.2} \\
            \midrule
            Paola & 0.17 & Emi-Paola & Kevin-Paola & \xmark & \textbf{0.2} & 0.78 & 0.68 & \textbf{0.59} & \textbf{0.8} & \textbf{0.77} & \textbf{0.05} & 0.87 \\
            & & & & \cmark & \textbf{0.2} & 0.78 & \textbf{0.71} & 0.52 & \textbf{0.8} & \textbf{0.77} & \textbf{0.05} & \textbf{1.05} \\
            \midrule
            Vanessa & 0.18 & Nebraas-Vanessa & Vanessa-Nicolas & \xmark & 0.11 & \textbf{0.8} & 0.68 & 0.43 & 0.44 & 0.34 & \textbf{0.08} & 1.9 \\
            & & & & \cmark & 0.12 & \textbf{0.8} & \textbf{0.82} & \textbf{0.63} & \textbf{0.45} & 0.33 & 0.09 & \textbf{1.22} \\
            \midrule
            Elise & 0.14 & Kevin-Elise & Emi-Elise & \xmark & 0.11 & 0.57 & 0.54 & 0.29 & 0.46 & 0.43 & \textbf{0.06} & 1.11 \\
            & & & & \cmark & \textbf{0.14} & \textbf{0.77} & \textbf{0.84} & \textbf{0.5} & \textbf{0.61} & \textbf{0.58} & 0.08 & \textbf{1.33} \\
            \midrule
            Fahim Khan & 0.12 & Fahim-Muhhamed & Fahim-Akib & \xmark & 0.1 & \textbf{0.73} & 0.42 & 0.29 & 0.36 & 0.32 & \textbf{0.05} & 2.35 \\
            & & & & \cmark & 0.1 & \textbf{0.73} & \textbf{0.71} & \textbf{0.59} & \textbf{0.43} & \textbf{0.35} & 0.06 & \textbf{1.1} \\
            \bottomrule
        \end{tabular}
    }
    \caption{\textbf{Content similarity and message-level EI comparison in persona simulation} for all speakers, with and without fine-tuning.}
    \label{tab:full-persona-simulation}
\end{table*}

\subsection{Context Experiments for Persona Simulation}
\label{appendix:persona-simulation-context}

\begin{figure}[!h]
    \centering
    \setlength{\tabcolsep}{0pt} 
    \includegraphics[width=0.9\columnwidth]{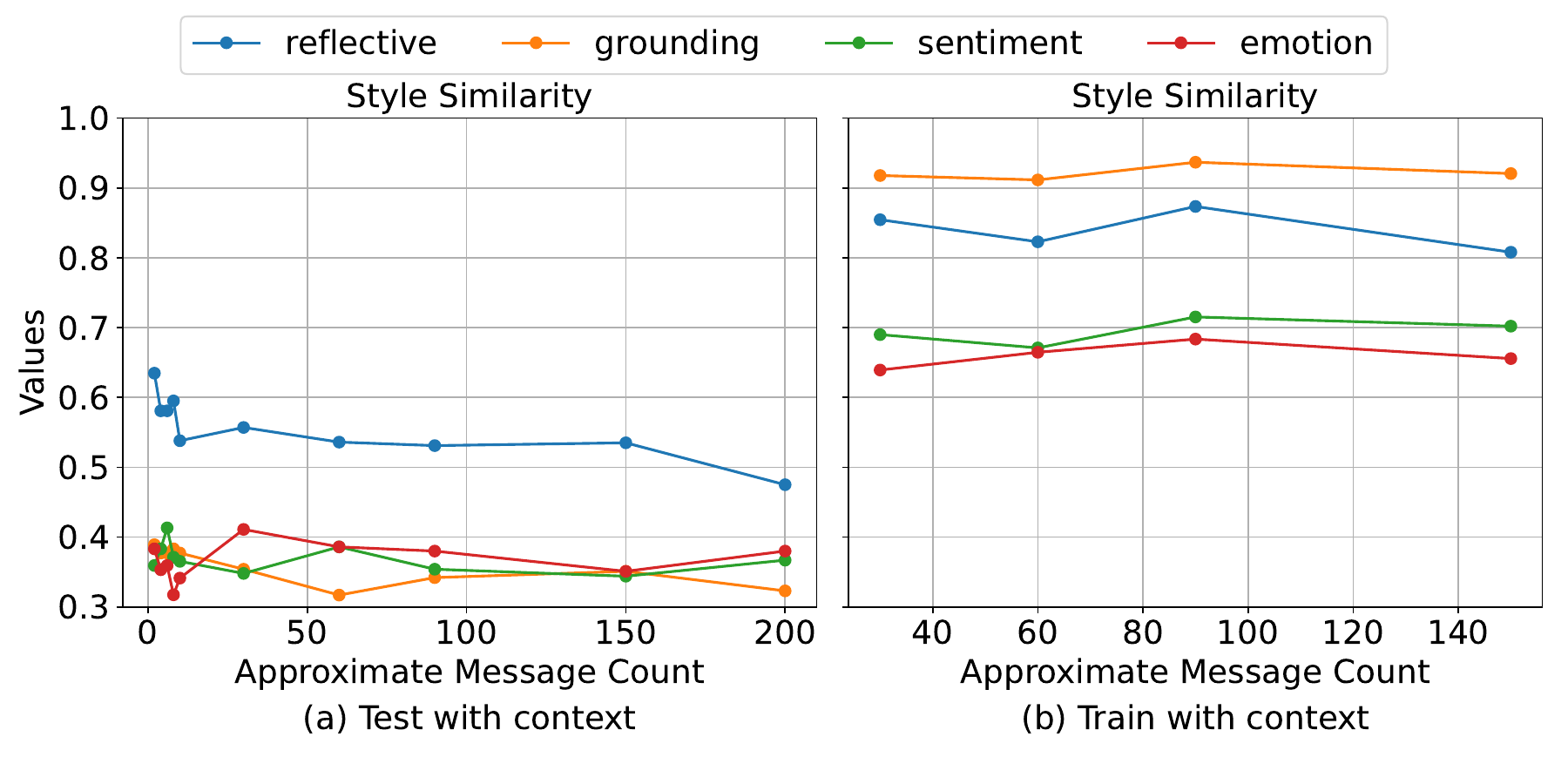}

    \vspace{-0.3cm}
    \caption{\textbf{Message-Level EI} in simulation for speaker ``Akib'' across different amounts of conversation history.  
    \textbf{(Left)} Zero-shot performance with varying conversation history lengths.  
    \textbf{(Right)} Fine-tuned performance with different amounts of conversation history.  
    Each point represents a model trained on a specific number of sessions and tested on the same number.}  

    \vspace{-0.3cm}
    \label{fig:simulation-by-context}
\end{figure}

\paragraph{Impact of conversational context.}  
We analyze how performance changes with varying amounts of provided context.  
Figure~\ref{fig:simulation-by-context} (left) shows that increasing conversation history does not improve message-level EI or exhibit a clear pattern.  
These findings suggest that LLMs struggle to capture and replicate a speaker’s style using context alone.  

\paragraph{Impact of fine-tuning.}  
We examine whether fine-tuning enhances an LLM’s ability to simulate a speaker’s responses given conversational context.  
Models are trained using different amounts of conversation history and tested with the same number of sessions used in training.  
Figure~\ref{fig:simulation-by-context} (right) shows that fine-tuning improves performance compared to the non-finetuned model.  
However, there is no clear trend indicating that increasing context length leads to further improvements.  
Performance appears to saturate after three sessions, suggesting that additional context beyond this point provides minimal benefit.  

\end{document}